\documentclass{article}
\usepackage{arxiv}
\usepackage[utf8]{inputenc} 
\usepackage[T1]{fontenc}    
\usepackage{hyperref}       
\usepackage{url}            
\usepackage{booktabs}       
\usepackage{amsfonts}       
\usepackage{nicefrac}       
\usepackage{microtype}      
\usepackage{lipsum}		    
\usepackage{graphicx}
\usepackage{doi}
\usepackage{glossaries}
\usepackage[table]{xcolor}
\definecolor{green}{rgb}{0,.5,0}
\definecolor{magenta}{rgb}{.75,0,.75}
\usepackage{soul}
\sethlcolor{lightgray}
\usepackage[labelfont=bf,font=footnotesize]{caption}
\usepackage{float}
\usepackage{multicol}
\usepackage{wrapfig} 
\usepackage{csquotes} 
\usepackage{tikz}
\newcommand{\review}[1]{\textcolor{black}{#1}}

\usepackage[breakable]{tcolorbox}
\usepackage{fancyvrb} 

\providecommand{\tightlist}{\setlength{\itemsep}{0pt}\setlength{\parskip}{0pt}}
\DefineVerbatimEnvironment{Highlighting}{Verbatim}{commandchars=\\\{\}}
\makeatletter
\def\PY@reset{\let\PY@it=\relax \let\PY@bf=\relax%
\let\PY@ul=\relax \let\PY@tc=\relax%
\let\PY@bc=\relax \let\PY@ff=\relax}
\def\PY@tok#1{\csname PY@tok@#1\endcsname}
\def\PY@toks#1+{\ifx\relax#1\empty\else%
\PY@tok{#1}\expandafter\PY@toks\fi}
\def\PY@do#1{\PY@bc{\PY@tc{\PY@ul{%
\PY@it{\PY@bf{\PY@ff{#1}}}}}}}
\def\PY#1#2{\PY@reset\PY@toks#1+\relax+\PY@do{#2}}
\@namedef{PY@tok@w}{\def\PY@tc##1{\textcolor[rgb]{0.73,0.73,0.73}{##1}}}
\@namedef{PY@tok@c}{\let\PY@it=\textit\def\PY@tc##1{\textcolor[rgb]{0.25,0.50,0.50}{##1}}}
\@namedef{PY@tok@cp}{\def\PY@tc##1{\textcolor[rgb]{0.74,0.48,0.00}{##1}}}
\@namedef{PY@tok@k}{\let\PY@bf=\textbf\def\PY@tc##1{\textcolor[rgb]{0.00,0.50,0.00}{##1}}}
\@namedef{PY@tok@kp}{\def\PY@tc##1{\textcolor[rgb]{0.00,0.50,0.00}{##1}}}
\@namedef{PY@tok@kt}{\def\PY@tc##1{\textcolor[rgb]{0.69,0.00,0.25}{##1}}}
\@namedef{PY@tok@o}{\def\PY@tc##1{\textcolor[rgb]{0.40,0.40,0.40}{##1}}}
\@namedef{PY@tok@ow}{\let\PY@bf=\textbf\def\PY@tc##1{\textcolor[rgb]{0.67,0.13,1.00}{##1}}}
\@namedef{PY@tok@nb}{\def\PY@tc##1{\textcolor[rgb]{0.00,0.50,0.00}{##1}}}
\@namedef{PY@tok@nf}{\def\PY@tc##1{\textcolor[rgb]{0.00,0.00,1.00}{##1}}}
\@namedef{PY@tok@nc}{\let\PY@bf=\textbf\def\PY@tc##1{\textcolor[rgb]{0.00,0.00,1.00}{##1}}}
\@namedef{PY@tok@nn}{\let\PY@bf=\textbf\def\PY@tc##1{\textcolor[rgb]{0.00,0.00,1.00}{##1}}}
\@namedef{PY@tok@ne}{\let\PY@bf=\textbf\def\PY@tc##1{\textcolor[rgb]{0.82,0.25,0.23}{##1}}}
\@namedef{PY@tok@nv}{\def\PY@tc##1{\textcolor[rgb]{0.10,0.09,0.49}{##1}}}
\@namedef{PY@tok@no}{\def\PY@tc##1{\textcolor[rgb]{0.53,0.00,0.00}{##1}}}
\@namedef{PY@tok@nl}{\def\PY@tc##1{\textcolor[rgb]{0.63,0.63,0.00}{##1}}}
\@namedef{PY@tok@ni}{\let\PY@bf=\textbf\def\PY@tc##1{\textcolor[rgb]{0.60,0.60,0.60}{##1}}}
\@namedef{PY@tok@na}{\def\PY@tc##1{\textcolor[rgb]{0.49,0.56,0.16}{##1}}}
\@namedef{PY@tok@nt}{\let\PY@bf=\textbf\def\PY@tc##1{\textcolor[rgb]{0.00,0.50,0.00}{##1}}}
\@namedef{PY@tok@nd}{\def\PY@tc##1{\textcolor[rgb]{0.67,0.13,1.00}{##1}}}
\@namedef{PY@tok@s}{\def\PY@tc##1{\textcolor[rgb]{0.73,0.13,0.13}{##1}}}
\@namedef{PY@tok@sd}{\let\PY@it=\textit\def\PY@tc##1{\textcolor[rgb]{0.73,0.13,0.13}{##1}}}
\@namedef{PY@tok@si}{\let\PY@bf=\textbf\def\PY@tc##1{\textcolor[rgb]{0.73,0.40,0.53}{##1}}}
\@namedef{PY@tok@se}{\let\PY@bf=\textbf\def\PY@tc##1{\textcolor[rgb]{0.73,0.40,0.13}{##1}}}
\@namedef{PY@tok@sr}{\def\PY@tc##1{\textcolor[rgb]{0.73,0.40,0.53}{##1}}}
\@namedef{PY@tok@ss}{\def\PY@tc##1{\textcolor[rgb]{0.10,0.09,0.49}{##1}}}
\@namedef{PY@tok@sx}{\def\PY@tc##1{\textcolor[rgb]{0.00,0.50,0.00}{##1}}}
\@namedef{PY@tok@m}{\def\PY@tc##1{\textcolor[rgb]{0.40,0.40,0.40}{##1}}}
\@namedef{PY@tok@gh}{\let\PY@bf=\textbf\def\PY@tc##1{\textcolor[rgb]{0.00,0.00,0.50}{##1}}}
\@namedef{PY@tok@gu}{\let\PY@bf=\textbf\def\PY@tc##1{\textcolor[rgb]{0.50,0.00,0.50}{##1}}}
\@namedef{PY@tok@gd}{\def\PY@tc##1{\textcolor[rgb]{0.63,0.00,0.00}{##1}}}
\@namedef{PY@tok@gi}{\def\PY@tc##1{\textcolor[rgb]{0.00,0.63,0.00}{##1}}}
\@namedef{PY@tok@gr}{\def\PY@tc##1{\textcolor[rgb]{1.00,0.00,0.00}{##1}}}
\@namedef{PY@tok@ge}{\let\PY@it=\textit}
\@namedef{PY@tok@gs}{\let\PY@bf=\textbf}
\@namedef{PY@tok@gp}{\let\PY@bf=\textbf\def\PY@tc##1{\textcolor[rgb]{0.00,0.00,0.50}{##1}}}
\@namedef{PY@tok@go}{\def\PY@tc##1{\textcolor[rgb]{0.53,0.53,0.53}{##1}}}
\@namedef{PY@tok@gt}{\def\PY@tc##1{\textcolor[rgb]{0.00,0.27,0.87}{##1}}}
\@namedef{PY@tok@err}{\def\PY@bc##1{{\setlength{\fboxsep}{\string -\fboxrule}\fcolorbox[rgb]{1.00,0.00,0.00}{1,1,1}{\strut ##1}}}}
\@namedef{PY@tok@kc}{\let\PY@bf=\textbf\def\PY@tc##1{\textcolor[rgb]{0.00,0.50,0.00}{##1}}}
\@namedef{PY@tok@kd}{\let\PY@bf=\textbf\def\PY@tc##1{\textcolor[rgb]{0.00,0.50,0.00}{##1}}}
\@namedef{PY@tok@kn}{\let\PY@bf=\textbf\def\PY@tc##1{\textcolor[rgb]{0.00,0.50,0.00}{##1}}}
\@namedef{PY@tok@kr}{\let\PY@bf=\textbf\def\PY@tc##1{\textcolor[rgb]{0.00,0.50,0.00}{##1}}}
\@namedef{PY@tok@bp}{\def\PY@tc##1{\textcolor[rgb]{0.00,0.50,0.00}{##1}}}
\@namedef{PY@tok@fm}{\def\PY@tc##1{\textcolor[rgb]{0.00,0.00,1.00}{##1}}}
\@namedef{PY@tok@vc}{\def\PY@tc##1{\textcolor[rgb]{0.10,0.09,0.49}{##1}}}
\@namedef{PY@tok@vg}{\def\PY@tc##1{\textcolor[rgb]{0.10,0.09,0.49}{##1}}}
\@namedef{PY@tok@vi}{\def\PY@tc##1{\textcolor[rgb]{0.10,0.09,0.49}{##1}}}
\@namedef{PY@tok@vm}{\def\PY@tc##1{\textcolor[rgb]{0.10,0.09,0.49}{##1}}}
\@namedef{PY@tok@sa}{\def\PY@tc##1{\textcolor[rgb]{0.73,0.13,0.13}{##1}}}
\@namedef{PY@tok@sb}{\def\PY@tc##1{\textcolor[rgb]{0.73,0.13,0.13}{##1}}}
\@namedef{PY@tok@sc}{\def\PY@tc##1{\textcolor[rgb]{0.73,0.13,0.13}{##1}}}
\@namedef{PY@tok@dl}{\def\PY@tc##1{\textcolor[rgb]{0.73,0.13,0.13}{##1}}}
\@namedef{PY@tok@s2}{\def\PY@tc##1{\textcolor[rgb]{0.73,0.13,0.13}{##1}}}
\@namedef{PY@tok@sh}{\def\PY@tc##1{\textcolor[rgb]{0.73,0.13,0.13}{##1}}}
\@namedef{PY@tok@s1}{\def\PY@tc##1{\textcolor[rgb]{0.73,0.13,0.13}{##1}}}
\@namedef{PY@tok@mb}{\def\PY@tc##1{\textcolor[rgb]{0.40,0.40,0.40}{##1}}}
\@namedef{PY@tok@mf}{\def\PY@tc##1{\textcolor[rgb]{0.40,0.40,0.40}{##1}}}
\@namedef{PY@tok@mh}{\def\PY@tc##1{\textcolor[rgb]{0.40,0.40,0.40}{##1}}}
\@namedef{PY@tok@mi}{\def\PY@tc##1{\textcolor[rgb]{0.40,0.40,0.40}{##1}}}
\@namedef{PY@tok@il}{\def\PY@tc##1{\textcolor[rgb]{0.40,0.40,0.40}{##1}}}
\@namedef{PY@tok@mo}{\def\PY@tc##1{\textcolor[rgb]{0.40,0.40,0.40}{##1}}}
\@namedef{PY@tok@ch}{\let\PY@it=\textit\def\PY@tc##1{\textcolor[rgb]{0.25,0.50,0.50}{##1}}}
\@namedef{PY@tok@cm}{\let\PY@it=\textit\def\PY@tc##1{\textcolor[rgb]{0.25,0.50,0.50}{##1}}}
\@namedef{PY@tok@cpf}{\let\PY@it=\textit\def\PY@tc##1{\textcolor[rgb]{0.25,0.50,0.50}{##1}}}
\@namedef{PY@tok@c1}{\let\PY@it=\textit\def\PY@tc##1{\textcolor[rgb]{0.25,0.50,0.50}{##1}}}
\@namedef{PY@tok@cs}{\let\PY@it=\textit\def\PY@tc##1{\textcolor[rgb]{0.25,0.50,0.50}{##1}}}
\def\PYZbs{\char`\\}
\def\PYZus{\char`\_}
\def\PYZob{\char`\{}
\def\PYZcb{\char`\}}

\def\PYZsh{\char`\#}

\def\PYZhy{\char`\-}

\def\PYZdq{\char`\"}

\makeatother
\makeatletter
    \newbox\Wrappedcontinuationbox 
    \newbox\Wrappedvisiblespacebox 
    \newcommand*\Wrappedvisiblespace {\textcolor{red}{\textvisiblespace}} 
    \newcommand*\Wrappedcontinuationsymbol {\textcolor{red}{\llap{\tiny$\m@th\hookrightarrow$}}} 
    \newcommand*\Wrappedcontinuationindent {3ex } 
    \newcommand*\Wrappedafterbreak {\kern\Wrappedcontinuationindent\copy\Wrappedcontinuationbox} 
    \newcommand*\Wrappedbreaksatspecials {%
        \def\PYGZus{\discretionary{\char`\_}{\Wrappedafterbreak}{\char`\_}}%
        \def\PYGZob{\discretionary{}{\Wrappedafterbreak\char`\{}{\char`\{}}%
        \def\PYGZcb{\discretionary{\char`\}}{\Wrappedafterbreak}{\char`\}}}%
        \def\PYGZca{\discretionary{\char`\^}{\Wrappedafterbreak}{\char`\^}}%
        \def\PYGZam{\discretionary{\char`\&}{\Wrappedafterbreak}{\char`\&}}%
        \def\PYGZlt{\discretionary{}{\Wrappedafterbreak\char`\<}{\char`\<}}%
        \def\PYGZgt{\discretionary{\char`\>}{\Wrappedafterbreak}{\char`\>}}%
        \def\PYGZsh{\discretionary{}{\Wrappedafterbreak\char`\#}{\char`\#}}%
        \def\PYGZpc{\discretionary{}{\Wrappedafterbreak\char`\%}{\char`\%}}%
        \def\PYGZdl{\discretionary{}{\Wrappedafterbreak\char`\$}{\char`\$}}%
        \def\PYGZhy{\discretionary{\char`\-}{\Wrappedafterbreak}{\char`\-}}%
        \def\PYGZsq{\discretionary{}{\Wrappedafterbreak\textquotesingle}{\textquotesingle}}%
        \def\PYGZdq{\discretionary{}{\Wrappedafterbreak\char`\"}{\char`\"}}%
        \def\PYGZti{\discretionary{\char`\~}{\Wrappedafterbreak}{\char`\~}}%
    } 
    \newcommand*\Wrappedbreaksatpunct {%
        \lccode`\~`\.\lowercase{\def~}{\discretionary{\hbox{\char`\.}}{\Wrappedafterbreak}{\hbox{\char`\.}}}%
        \lccode`\~`\,\lowercase{\def~}{\discretionary{\hbox{\char`\,}}{\Wrappedafterbreak}{\hbox{\char`\,}}}%
        \lccode`\~`\;\lowercase{\def~}{\discretionary{\hbox{\char`\;}}{\Wrappedafterbreak}{\hbox{\char`\;}}}%
        \lccode`\~`\:\lowercase{\def~}{\discretionary{\hbox{\char`\:}}{\Wrappedafterbreak}{\hbox{\char`\:}}}%
        \lccode`\~`\?\lowercase{\def~}{\discretionary{\hbox{\char`\?}}{\Wrappedafterbreak}{\hbox{\char`\?}}}%
        \lccode`\~`\!\lowercase{\def~}{\discretionary{\hbox{\char`\!}}{\Wrappedafterbreak}{\hbox{\char`\!}}}%
        \lccode`\~`\/\lowercase{\def~}{\discretionary{\hbox{\char`\/}}{\Wrappedafterbreak}{\hbox{\char`\/}}}%
        \catcode`\.\active
        \catcode`\,\active 
        \catcode`\;\active
        \catcode`\:\active
        \catcode`\?\active
        \catcode`\!\active
        \catcode`\/\active 
        \lccode`\~`\~ 	
    }
\makeatother
\let\OriginalVerbatim=\Verbatim
\makeatletter
\renewcommand{\Verbatim}[1][1]{%
    \sbox\Wrappedcontinuationbox {\Wrappedcontinuationsymbol}%
    \sbox\Wrappedvisiblespacebox {\FV@SetupFont\Wrappedvisiblespace}%
    \def\FancyVerbFormatLine ##1{\hsize\linewidth
        \vtop{\raggedright\hyphenpenalty\z@\exhyphenpenalty\z@
            \doublehyphendemerits\z@\finalhyphendemerits\z@
            \strut ##1\strut}%
    }%
    \def\FV@Space {%
        \nobreak\hskip\z@ plus\fontdimen3\font minus\fontdimen4\font
        \discretionary{\copy\Wrappedvisiblespacebox}{\Wrappedafterbreak}
        {\kern\fontdimen2\font}%
    }%
    
    \Wrappedbreaksatspecials
    \OriginalVerbatim[#1,codes*=\Wrappedbreaksatpunct]%
}
\makeatother
\definecolor{incolor}{HTML}{303F9F}
\definecolor{outcolor}{HTML}{D84315}
\definecolor{cellborder}{HTML}{CFCFCF}
\definecolor{cellbackground}{HTML}{F7F7F7}
\makeatletter
\newcommand{\boxspacing}{\kern\kvtcb@left@rule\kern\kvtcb@boxsep}
\makeatother
\newcommand{\prompt}[4]{{\ttfamily\llap{{\color{#2}[#3]:\hspace{3pt}#4}}\vspace{-\baselineskip}}}
\title{DeepVoxNet2: Yet another CNN framework}
\renewcommand{\undertitle}{From pre- to postprocessing and keeping track of the spatial origin of the data}
\renewcommand{\headeright}{}
\renewcommand{\shorttitle}{DeepVoxNet2: Yet another CNN framework}
\author{
    \href{https://orcid.org/0000-0001-7206-2671}{\includegraphics[scale=0.06]{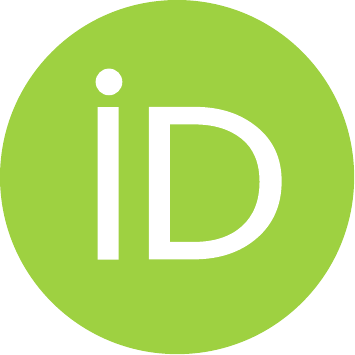}\hspace{1mm}Jeroen Bertels}\\
	Processing Speech and Images\\
	Department of Electrical Engineering\\
	KU Leuven, Belgium\\
	\texttt{jeroen.bertels@kuleuven.be}\\
	\And
	David Robben\\
	Processing Speech and Images\\
	Department of Electrical Engineering\\
	KU Leuven, Belgium\\
	\texttt{david.robben@kuleuven.be}\\
    \And
	Robin Lemmens\\
	Laboratory of Neurobiology\\
	Department of Neurosciences\\
	KU Leuven, Belgium\\
	\texttt{robin.lemmens@kuleuven.be}\\
	\And
	Dirk Vandermeulen\\
	Processing Speech and Images\\
	Department of Electrical Engineering\\
	KU Leuven, Belgium\\
	\texttt{dirk.vandermeulen@kuleuven.be}
}
\begin{document}
\date{}  
\maketitle
\newglossaryentry{cbf}{name={CBF},description={cerebral blood flow}}
\newglossaryentry{cbv}{name={CBV},description={cerebral blood volume}}
\newglossaryentry{cpp}{name={CPP},description={cerebral perfusion pressure}}
\newglossaryentry{cvr}{name={CVR},description={cerebrovascular resistance}}
\newglossaryentry{oef}{name={OEF},description={oxygen extraction fraction}}
\newglossaryentry{ais}{name={AIS},description={acute ischemic stroke}}
\newglossaryentry{tso}{name={TsO},description={time since onset}}
\newglossaryentry{ttt}{name={TtT},description={time to treatment}}
\newglossaryentry{tici}{name={mTICI},description={modified Thrombolysis in Cerebral Infarction}}
\newglossaryentry{ivt}{name={IVT},description={intravenous treatment}}
\newglossaryentry{iat}{name={IAT},description={intra-arterial treatment}}
\newglossaryentry{evt}{name={EVT},description={endovascular treatment}}
\newglossaryentry{mrs}{name={mRS},description={modified Rankin Scale}}
\newglossaryentry{nihss}{name={NIHSS},description={National Institute of Health Stroke Scale}}
\newglossaryentry{ich}{name={ICH},description={intracranial hemorrhage}}
\newglossaryentry{ct}{name={CT},description={computed tomography}}
\newglossaryentry{ncct}{name={NCCT},description={non-enhanced or non-contrast CT}}
\newglossaryentry{aspects}{name={ASPECTS},description={Alberta Stroke Program Early CT Score}}
\newglossaryentry{mca}{name={MCA},description={middle cerebral artery}}
\newglossaryentry{mri}{name={MRI},description={magnetic resonance imaging}}
\newglossaryentry{dwi}{name={DWI},description={diffusion weighted imaging}}
\newglossaryentry{adc}{name={ADC},description={apparent diffusion coefficient}}
\newglossaryentry{ctp}{name={CTP},description={CT perfusion}}
\newglossaryentry{cta}{name={CTA},description={CT angiography}}
\newglossaryentry{pwi}{name={PWI},description={perfusion-weighted imaging}}
\newglossaryentry{tcc}{name={TCC},description={time concentration curve}}
\newglossaryentry{irf}{name={irf},description={impulse response function}}
\newglossaryentry{imp}{name={imp},description={Dirac impulse}}
\newglossaryentry{aif}{name={AIF},description={arterial imput function}}
\newglossaryentry{pet}{name={PET},description={positron emission tomography}}
\newglossaryentry{spect}{name={SPECT},description={single photon emission CT}}
\newglossaryentry{mra}{name={MRA},description={MRI angiography}}
\newglossaryentry{cc}{name={CC},description={collateral circulation}}
\newglossaryentry{3d}{name={3D},description={3 dimensional}}
\newglossaryentry{alara}{name={ALARA},description={as low as reasonably possible}}
\newglossaryentry{dcv}{name={DCV},description={deconvolution}}
\newglossaryentry{cnn}{name={CNN},description={convolutional neural network}}
\newglossaryentry{tmax}{name={Tmax},description={location of maximum of TCC\textsubscript{irf}}}
\newglossaryentry{rcbf}{name={rCBF},description={relative CBF}}
\newglossaryentry{cbct}{name={CBCT},description={C-arm or cone-beam CT}}
\newglossaryentry{cbctp}{name={CBCTP},description={CBCT perfusion}}
\newglossaryentry{de}{name={DE},description={dual-energy}}
\newglossaryentry{dencct}{name={DENCCT},description={DE NCCT}}
\newglossaryentry{rcbv}{name={rCBV},description={relative CBV}}
\newglossaryentry{o24h}{name={O24h},description={occlusion present at follow-up}}
\newglossaryentry{flair}{name={FLAIR},description={fluid-attenuated inversion recovery MRI}}
\newglossaryentry{vof}{name={VOF},description={venous output function}}
\newglossaryentry{glm}{name={GLM},description={generalized linear model}}
\newglossaryentry{csf}{name={CSF},description={cerebrospinal fluid}}
\newglossaryentry{minip}{name={minIP},description={minimum intensity projection}}
\newglossaryentry{meanip}{name={meanIP},description={mean intensity projection}}
\newglossaryentry{maxip}{name={maxIP},description={maximum intensity projection}}
\newglossaryentry{relu}{name={ReLU},description={rectified linear unit}}
\newglossaryentry{ica}{name={ICA},description={internal carotid artery}}
\newglossaryentry{aca}{name={ACA},description={arterior cerebral artery}}
\newglossaryentry{rep+}{name={Rep$^+$},description={(the group of) reperfusers}}
\newglossaryentry{rep-}{name={Rep$^-$},description={(the group of) non-reperfusers}}
\newglossaryentry{rep+-}{name={Rep$^{+/-}$},description={combination of Rep$^+$ and Rep$^-$}}
\newglossaryentry{lps}{name={LPS},description={left posterior superior}}
\newglossaryentry{mse}{name={MSE},description={mean squared error}}
\newglossaryentry{adv}{name={$|\Delta\mathrm{V}|$},description={absolute volume error}}
\newglossaryentry{rf}{name={RF},description={receptive field}}
\newglossaryentry{prf}{name={pRF},description={physical receptive field}}
\newglossaryentry{ce}{name={CE},description={cross-entropy}}
\newglossaryentry{sgd}{name={SGD},description={stochastic gradient descent}}
\newglossaryentry{dsc}{name={DSC},description={Dice coefficient}}
\newglossaryentry{dv}{name={$\Delta\mathrm{V}$},description={volume bias or error}}
\newglossaryentry{hd95}{name={HD95},description={95th percentile Hausdorff distance}}
\newglossaryentry{ppv}{name={PPV},description={precision or positive predictive value}}
\newglossaryentry{tpr}{name={TPR},description={recall or true positive rate}}
\newglossaryentry{ece}{name={ECE},description={expected calibration error}}
\newglossaryentry{auc}{name={AUC},description={area under the precision-recall curve}}
\newglossaryentry{fov}{name={FOV},description={field of view}}
\newglossaryentry{m}{name={M},description={MRCLEAN}}
\newglossaryentry{c}{name={C},description={CRISP}}
\newglossaryentry{k}{name={K},description={KAROLINSKA}}
\newglossaryentry{mck}{name={MCK},description={combination of M, C and K}}
\newglossaryentry{clpr}{name={ClPr},description={clinical practice}}
\newglossaryentry{jl}{name={JL},description={Julie Lambert}}
\newglossaryentry{jd}{name={JD},description={Jelle Demeestere}}
\newglossaryentry{grt}{name={GrT},description={grow time}}
\newglossaryentry{bao}{name={BAO},description={basilar artery occlusion}}
\newglossaryentry{rtpa}{name={rtPA},description={recombinant tissue plasminogen activator}}
\newglossaryentry{nn}{name={NN},description={neural network}}
\newglossaryentry{gpu}{name={GPU},description={graphics processing unit}}
\newglossaryentry{rgb}{name={RGB},description={red green blue}}
\newglossaryentry{pca}{name={PCA},description={principle component analysis}}
\newglossaryentry{fcn}{name={FCN},description={fully-convolutional network}}
\newglossaryentry{dl}{name={DL},description={Dice loss}}
\newglossaryentry{sd}{name={SD},description={soft Dice}}
\newglossaryentry{gan}{name={GAN},description={generative adversarial network}}
\newglossaryentry{t1w}{name={T1w},description={T1-weighted MRI}}
\newglossaryentry{stn}{name={STN},description={spatial transformer network}}
\newglossaryentry{dvn2}{name={DVN2},description={DeepVoxNet2}}
\newglossaryentry{gt}{name={GT},description={ground truth}}
\newglossaryentry{vs}{name={VS},description={voxel size}}
\newglossaryentry{lvo}{name={LVO},description={large vessel occlusion}}
\newglossaryentry{nexis}{name={NEXIS},description={NExt generation X-ray Imaging System}}
\newglossaryentry{ip}{name={IP},description={intensity projection}}
\begin{abstract}
	We know that both the \gls{cnn} mapping function and the sampling scheme are of paramount importance for \gls{cnn}-based image analysis. It is clear that both functions operate in the same space, with an image axis $\mathcal{I}$ and a feature axis $\mathcal{F}$. Remarkably, we found that no frameworks existed that unified the two and kept track of the spatial origin of the data automatically. Based on our own practical experience, we found the latter to often result in complex coding and pipelines that are difficult to exchange. This article introduces our framework for 1, 2 or 3D image classification or segmentation: DeepVoxNet2 (\gls{dvn2}). This article serves as an interactive tutorial, and a pre-compiled version, including the outputs of the code blocks, can be found online in the public \gls{dvn2} repository. This tutorial uses data from the multimodal Brain Tumor Image Segmentation Benchmark (BRATS) of 2018 to show an example of a 3D segmentation pipeline.
\end{abstract}
    \section{Introduction}\label{introduction}
Yes, we are introducing yet another \gls{cnn} framework. However, \gls{dvn2} also includes many other essential tools to train and test a \gls{cnn} but also to process medical images, organize the experiments and analyze the results. This tutorial uses data from the multimodal Brain Tumor Image Segmentation Benchmark (BRATS) of 2018~\cite{Menze2015,Bakas2017,Bakas2018} to show an example of a 3D segmentation pipeline.
\subsection{Installation}\label{installation}
The \gls{dvn2} repository~\cite{deepvoxnet2} can be installed as a Python library such that all
dependencies are also configured correctly. Even when you do
not want to use the functionalities of \gls{dvn2} explicitly, installing it
will provide users with a complete and working Python environment for
doing medical image analysis in general.
As of now, it is best to add
\gls{dvn2} to an empty Python 3.9 environment, e.g.~using the Anaconda package
manager to first create and activate such an environment via:

\begin{verbatim}
conda create --name dvn2env python=3.9
conda activate dvn2env
\end{verbatim}

Once you have activated your Python 3.9 environment correctly, there are
two options to install \gls{dvn2} together with all its dependencies, and, as a
result, make your environment medical image analysis proof:

\begin{itemize}
\tightlist
\item
  First cloning or downloading the repository and then via:
\end{itemize}

\begin{verbatim}
pip install -e /path/to/deepvoxnet2
\end{verbatim}

\begin{itemize}
\tightlist
\item
  Installing it directly from Github via:
\end{itemize}

\begin{verbatim}
pip install git+https://github.com/JeroenBertels/deepvoxnet2
\end{verbatim}

The first method is helpful if you want to develop and
contribute to the \gls{dvn2} library. To upgrade your installation
using the first method, download the latest version and repeat
the process or \texttt{git\ pull} the new version. Repeat the command when using the second
method but add the \texttt{-\/upgrade} flag.
You can also install or revert to a specific version of \gls{dvn2}. In that
case, append \texttt{@version\_tag} (e.g., ~\emph{@deepvoxnet-2.12.1}) to
the paths in the above commands. Some functions require the SimpleITK
and SimpleElastix software to be installed. To install these packages,
please append the paths in the above commands with \texttt{{[}sitk{]}}.
These are not installed by default because they are only available for a
limited number of operating systems.

\subsection{Schematic overview}\label{schematic-overview}

The main driver behind the creation of \gls{dvn2} was that we wanted to make something that gives insights rather than being super efficient or fully self-configuring. To the best of our knowledge, most libraries are oriented towards the latter, and there are no libraries that follow the supervised learning paradigm more closely. In \gls{dvn2}, the focus is on transforming one set of \emph{data pairs} into another.  We make no difference between the CNN and sampling function, seeing them as mapping functions that transform one distribution into another, albeit implicitly, and keep track of the spatial aspect of the data. By doing so, the entire pipeline can be constructed in an end-to-end fashion:

\begin{equation}
    \mathcal{\hat{S}'} \xleftarrow[]{put} \mathcal{\hat{S}} \xleftarrow[]{m} \mathcal{S} \xleftarrow[]{s''} \mathcal{S'} \xleftarrow[]{s'} P(X,Y).\label{eq:dvn_eq}
\end{equation}

With $P(X,Y)$, we denote the joint probability
distribution of all the \emph{elements} that make up a data pair and is
often present only implicitly as the empirical distribution arising from
the observed set of data pairs \(\mathcal{S'}\). In a straightforward supervised learning setting, a data pair may consist of
the input variable $X$ and output variable $Y$. However, in \gls{dvn2} we
generalized this concept and made no difference between $X$ and $Y$. They can be represented using the same abstraction, i.e.~the
Sample object (Section~\ref{the-sample-object}). Furthermore, a data pair does not necessarily
represent a pair but can represent any number of Sample objects. To
organize a data pair and the set of data pairs \(\mathcal{S'}\) in a
structured manner, in \gls{dvn2} we can use the hierarchical structure of the
Mirc, Dataset, Case, Record and Modality objects (Section~\ref{organizing-the-data}).

In \gls{dvn2}, we represent any transformed set \(\mathcal{S}\) implicitly,
by implementing the sampling function \(s''\) using the successive
application of (i) a Sampler object that can sample a data pair from
\(\mathcal{S'}\) in the form of an Identifier object (Section~\ref{the-identifier-object}), and (ii)
a network of Transformer objects (as a Creator object) that can
transform the data pair (Section~\ref{data-transformation}). A crucial design aspect is that the
Transformer keeps track of the spatiality of a Sample object, thus
keeping track of the voxel-to-world transformation matrix. The
combination of this, together with the fact that a Transformer is a
generator, allows the construction of complex pipelines end-to-end. This
way, we can view a \gls{cnn} also as a Transformer that converts a data pair
into another.

By viewing the \gls{cnn} as just another Transformer, we were able to create
the holistic \gls{cnn} framework that \gls{dvn2} is (Section~\ref{dvn2-for-cnn-based-applications}). If, for example,
\(\mathcal{S'}\) operates in the original image space and the set
\(\mathcal{S}\) uses image crops, you can extend the Transformer
network straightforwardly.
For example, first resulting in a set \(\mathcal{\hat{S}}\) after application of the CNN, e.g., as the sampling function $m$, and then putting back the patches such that  \(\mathcal{\hat{S}'}\) operates again in the original image space. As a result, we can build
one large Transformer network and select which outputs are used
for training the \gls{cnn}.

Furthermore, \gls{dvn2} contains functions to construct U-Net and DeepMedic
\gls{cnn} architectures and gives insights about crucial design aspects such
as the \gls{rf} and the possible output sizes. It also contains various metrics and losses that consistently work and give
results in 5D. Other functionalities are related to analyses
(e.g.~statistical testing), conversions (e.g.~image I/O such as dicom
reading), transformations (e.g.~registration, resampling) and plotting
(Section~\ref{other-functionalities-in-dvn2}).

    \section{The Sample object}\label{the-sample-object}

One of the core objects of \gls{dvn2} is the Sample object. In line with the
definitions of \(X\) and \(Y\) from the previous chapter, the Sample
represents an array with a batch axis \(\mathcal{B}\), a spatial axis
\(\mathcal{I}\) and a feature axis \(\mathcal{F}\). The spatial axis
itself is 3D, hence the Sample object is a 5D array of size B x
\(\mathrm{I}_0\) x \(\mathrm{I}_1\) x \(\mathrm{I}_2\) x F. Different
from the standard Numpy array, the Sample object also has an affine
attribute. This affine attribute stores the voxel-to-world
transformation matrix for each batch element. As a result, the affine is
a 3D array of size B x 4 x 4, thus assuming the same voxel-to-world
transformation for each feature \(f\) given a batch element \(b\).

    \begin{tcolorbox}[breakable, size=fbox, boxrule=1pt, pad at break*=1mm,colback=cellbackground, colframe=cellborder]
\prompt{In}{incolor}{ }{\boxspacing}
\begin{Verbatim}[commandchars=\\\{\}]
\PY{k+kn}{import} \PY{n+nn}{numpy} \PY{k}{as} \PY{n+nn}{np}
\PY{k+kn}{from} \PY{n+nn}{deepvoxnet2}\PY{n+nn}{.}\PY{n+nn}{components}\PY{n+nn}{.}\PY{n+nn}{sample} \PY{k+kn}{import} \PY{n}{Sample}

\PY{n}{array\PYZus{}of\PYZus{}ones} \PY{o}{=} \PY{n}{np}\PY{o}{.}\PY{n}{ones}\PY{p}{(}\PY{p}{(}\PY{l+m+mi}{2}\PY{p}{,} \PY{l+m+mi}{100}\PY{p}{,} \PY{l+m+mi}{100}\PY{p}{,} \PY{l+m+mi}{3}\PY{p}{)}\PY{p}{)}
\PY{n}{sample\PYZus{}of\PYZus{}ones} \PY{o}{=} \PY{n}{Sample}\PY{p}{(}\PY{n}{array\PYZus{}of\PYZus{}ones}\PY{p}{,} \PY{n}{affine}\PY{o}{=}\PY{k+kc}{None}\PY{p}{)}
\PY{n+nb}{print}\PY{p}{(}\PY{n}{sample\PYZus{}of\PYZus{}ones}\PY{o}{.}\PY{n}{shape}\PY{p}{)}
\PY{n+nb}{print}\PY{p}{(}\PY{n}{sample\PYZus{}of\PYZus{}ones}\PY{o}{.}\PY{n}{affine}\PY{p}{)}
\end{Verbatim}
\end{tcolorbox}

    It should be clear that, upon creation, the Sample object ensures the
underlying array is 5D and that an affine attribute is specified. Therefore, be
careful and specific. It is best practice always to provide the entire
5D array and affine attribute yourself to avoid unexpected behavior. In that
sense, when working with only two spatial dimensions, it
would be straightforward to set the fourth dimension of the Sample
object as the singleton dimension.

    \begin{tcolorbox}[breakable, size=fbox, boxrule=1pt, pad at break*=1mm,colback=cellbackground, colframe=cellborder]
\prompt{In}{incolor}{ }{\boxspacing}
\begin{Verbatim}[commandchars=\\\{\}]
\PY{n}{default\PYZus{}affine} \PY{o}{=} \PY{n}{np}\PY{o}{.}\PY{n}{eye}\PY{p}{(}\PY{l+m+mi}{4}\PY{p}{)}
\PY{n}{sample\PYZus{}of\PYZus{}ones} \PY{o}{=} \PY{n}{Sample}\PY{p}{(}\PY{n}{array\PYZus{}of\PYZus{}ones}\PY{p}{[}\PY{p}{:}\PY{p}{,} \PY{p}{:}\PY{p}{,} \PY{p}{:}\PY{p}{,} \PY{k+kc}{None}\PY{p}{,} \PY{p}{:}\PY{p}{]}\PY{p}{,} \PY{n}{affine}\PY{o}{=}\PY{n}{np}\PY{o}{.}\PY{n}{stack}\PY{p}{(}\PY{p}{[}\PY{n}{default\PYZus{}affine}\PY{p}{]} \PY{o}{*} \PY{l+m+mi}{2}\PY{p}{,} \PY{n}{axis}\PY{o}{=}\PY{l+m+mi}{0}\PY{p}{)}\PY{p}{)}
\PY{n+nb}{print}\PY{p}{(}\PY{n}{sample\PYZus{}of\PYZus{}ones}\PY{o}{.}\PY{n}{shape}\PY{p}{)}
\PY{n+nb}{print}\PY{p}{(}\PY{n}{sample\PYZus{}of\PYZus{}ones}\PY{o}{.}\PY{n}{affine}\PY{p}{)}
\end{Verbatim}
\end{tcolorbox}

    You may have a medical image, for example, in Nifti
format, and you want to convert this into a Sample object. In this
case, you can load the image and use the internal data and affine attribute to
create the Sample. Below we do this for the \gls{flair} image and the ground
truth, here, the whole tumor segmentation of the first subject. It should not
be surprising that both the \gls{flair} and ground truth are equal in size and have the
same voxel-to-word transformation.

    \begin{tcolorbox}[breakable, size=fbox, boxrule=1pt, pad at break*=1mm,colback=cellbackground, colframe=cellborder]
\prompt{In}{incolor}{ }{\boxspacing}
\begin{Verbatim}[commandchars=\\\{\}]
\PY{k+kn}{import} \PY{n+nn}{os}
\PY{k+kn}{import} \PY{n+nn}{nibabel} \PY{k}{as} \PY{n+nn}{nib}
\PY{k+kn}{from} \PY{n+nn}{deepvoxnet2} \PY{k+kn}{import} \PY{n}{DEMO\PYZus{}DIR}  \PY{c+c1}{\PYZsh{} modify the DEMO\PYZus{}DIR accordingly}

\PY{n}{flair\PYZus{}path} \PY{o}{=} \PY{n}{os}\PY{o}{.}\PY{n}{path}\PY{o}{.}\PY{n}{join}\PY{p}{(}\PY{n}{DEMO\PYZus{}DIR}\PY{p}{,} \PY{l+s+s2}{\PYZdq{}}\PY{l+s+s2}{brats\PYZus{}2018}\PY{l+s+s2}{\PYZdq{}}\PY{p}{,} \PY{l+s+s2}{\PYZdq{}}\PY{l+s+s2}{case\PYZus{}0}\PY{l+s+s2}{\PYZdq{}}\PY{p}{,} \PY{l+s+s2}{\PYZdq{}}\PY{l+s+s2}{FLAIR.nii.gz}\PY{l+s+s2}{\PYZdq{}}\PY{p}{)}
\PY{n}{flair\PYZus{}image} \PY{o}{=} \PY{n}{nib}\PY{o}{.}\PY{n}{load}\PY{p}{(}\PY{n}{flair\PYZus{}path}\PY{p}{)}
\PY{n+nb}{print}\PY{p}{(}\PY{n}{flair\PYZus{}image}\PY{o}{.}\PY{n}{shape}\PY{p}{)}
\PY{n+nb}{print}\PY{p}{(}\PY{n}{flair\PYZus{}image}\PY{o}{.}\PY{n}{affine}\PY{p}{)}
\PY{n}{flair\PYZus{}sample} \PY{o}{=} \PY{n}{Sample}\PY{p}{(}\PY{n}{flair\PYZus{}image}\PY{o}{.}\PY{n}{get\PYZus{}fdata}\PY{p}{(}\PY{p}{)}\PY{p}{[}\PY{k+kc}{None}\PY{p}{,} \PY{o}{.}\PY{o}{.}\PY{o}{.}\PY{p}{,} \PY{k+kc}{None}\PY{p}{]}\PY{p}{,} \PY{n}{affine}\PY{o}{=}\PY{n}{flair\PYZus{}image}\PY{o}{.}\PY{n}{affine}\PY{p}{[}\PY{k+kc}{None}\PY{p}{,} \PY{o}{.}\PY{o}{.}\PY{o}{.}\PY{p}{]}\PY{p}{)}
\PY{n+nb}{print}\PY{p}{(}\PY{n}{flair\PYZus{}sample}\PY{o}{.}\PY{n}{shape}\PY{p}{)}
\PY{n+nb}{print}\PY{p}{(}\PY{n}{flair\PYZus{}sample}\PY{o}{.}\PY{n}{affine}\PY{p}{)}
\PY{n}{gt\PYZus{}path} \PY{o}{=} \PY{n}{os}\PY{o}{.}\PY{n}{path}\PY{o}{.}\PY{n}{join}\PY{p}{(}\PY{n}{DEMO\PYZus{}DIR}\PY{p}{,} \PY{l+s+s2}{\PYZdq{}}\PY{l+s+s2}{brats\PYZus{}2018}\PY{l+s+s2}{\PYZdq{}}\PY{p}{,} \PY{l+s+s2}{\PYZdq{}}\PY{l+s+s2}{case\PYZus{}0}\PY{l+s+s2}{\PYZdq{}}\PY{p}{,} \PY{l+s+s2}{\PYZdq{}}\PY{l+s+s2}{GT\PYZus{}W.nii.gz}\PY{l+s+s2}{\PYZdq{}}\PY{p}{)}
\PY{n}{gt\PYZus{}image} \PY{o}{=} \PY{n}{nib}\PY{o}{.}\PY{n}{load}\PY{p}{(}\PY{n}{gt\PYZus{}path}\PY{p}{)}
\PY{n}{gt\PYZus{}sample} \PY{o}{=} \PY{n}{Sample}\PY{p}{(}\PY{n}{gt\PYZus{}image}\PY{o}{.}\PY{n}{get\PYZus{}fdata}\PY{p}{(}\PY{p}{)}\PY{p}{[}\PY{k+kc}{None}\PY{p}{,} \PY{o}{.}\PY{o}{.}\PY{o}{.}\PY{p}{,} \PY{k+kc}{None}\PY{p}{]}\PY{p}{,} \PY{n}{affine}\PY{o}{=}\PY{n}{gt\PYZus{}image}\PY{o}{.}\PY{n}{affine}\PY{p}{[}\PY{k+kc}{None}\PY{p}{,} \PY{o}{.}\PY{o}{.}\PY{o}{.}\PY{p}{]}\PY{p}{)}
\PY{n+nb}{print}\PY{p}{(}\PY{n}{gt\PYZus{}sample}\PY{o}{.}\PY{n}{shape}\PY{p}{)}
\PY{n+nb}{print}\PY{p}{(}\PY{n}{gt\PYZus{}sample}\PY{o}{.}\PY{n}{affine}\PY{p}{)}
\end{Verbatim}
\end{tcolorbox}

    It should be clear that a Sample object is not so different from a Numpy
array. If you do not care about the spatiality of the data, you can
always use the default affine and ignore it. Let us already define a
visualization function that will become handy later on in this tutorial.
For a list of Sample objects, it will simply depict the first batch
element and first feature the middle slice of each Sample in the list.

    \begin{tcolorbox}[breakable, size=fbox, boxrule=1pt, pad at break*=1mm,colback=cellbackground, colframe=cellborder]
\prompt{In}{incolor}{ }{\boxspacing}
\begin{Verbatim}[commandchars=\\\{\}]
\PY{k+kn}{from} \PY{n+nn}{matplotlib} \PY{k+kn}{import} \PY{n}{pyplot} \PY{k}{as} \PY{n}{plt}

\PY{k}{def} \PY{n+nf}{visualize\PYZus{}list\PYZus{}of\PYZus{}samples}\PY{p}{(}\PY{n}{list\PYZus{}of\PYZus{}samples}\PY{p}{)}\PY{p}{:}
    \PY{n}{n\PYZus{}cols} \PY{o}{=} \PY{n+nb}{len}\PY{p}{(}\PY{n}{list\PYZus{}of\PYZus{}samples}\PY{p}{)}
    \PY{n}{fig}\PY{p}{,} \PY{n}{axs} \PY{o}{=} \PY{n}{plt}\PY{o}{.}\PY{n}{subplots}\PY{p}{(}\PY{l+m+mi}{1}\PY{p}{,} \PY{n}{n\PYZus{}cols}\PY{p}{,} \PY{n}{figsize}\PY{o}{=}\PY{p}{(}\PY{n}{n\PYZus{}cols} \PY{o}{*} \PY{l+m+mi}{3} \PY{o}{*} \PY{n}{list\PYZus{}of\PYZus{}samples}\PY{p}{[}\PY{l+m+mi}{0}\PY{p}{]}\PY{o}{.}\PY{n}{shape}\PY{p}{[}\PY{l+m+mi}{1}\PY{p}{]} \PY{o}{/} \PY{l+m+mi}{240}\PY{p}{,} \PY{l+m+mi}{3} \PY{o}{*} \PY{n}{list\PYZus{}of\PYZus{}samples}\PY{p}{[}\PY{l+m+mi}{0}\PY{p}{]}\PY{o}{.}\PY{n}{shape}\PY{p}{[}\PY{l+m+mi}{1}\PY{p}{]} \PY{o}{/} \PY{l+m+mi}{240}\PY{p}{)}\PY{p}{)}
    \PY{n}{axs} \PY{o}{=} \PY{p}{[}\PY{n}{axs}\PY{p}{]} \PY{k}{if} \PY{n}{n\PYZus{}cols} \PY{o}{==} \PY{l+m+mi}{1} \PY{k}{else} \PY{n}{axs}
    \PY{k}{for} \PY{n}{i}\PY{p}{,} \PY{n}{sample} \PY{o+ow}{in} \PY{n+nb}{enumerate}\PY{p}{(}\PY{n}{list\PYZus{}of\PYZus{}samples}\PY{p}{)}\PY{p}{:}
        \PY{n}{axs}\PY{p}{[}\PY{n}{i}\PY{p}{]}\PY{o}{.}\PY{n}{imshow}\PY{p}{(}\PY{n}{sample}\PY{p}{[}\PY{l+m+mi}{0}\PY{p}{,} \PY{p}{:}\PY{p}{,} \PY{p}{:}\PY{p}{,} \PY{n}{sample}\PY{o}{.}\PY{n}{shape}\PY{p}{[}\PY{l+m+mi}{3}\PY{p}{]} \PY{o}{/}\PY{o}{/} \PY{l+m+mi}{2}\PY{p}{,} \PY{l+m+mi}{0}\PY{p}{]}\PY{o}{.}\PY{n}{T}\PY{p}{,} \PY{n}{cmap}\PY{o}{=}\PY{l+s+s2}{\PYZdq{}}\PY{l+s+s2}{gray}\PY{l+s+s2}{\PYZdq{}}\PY{p}{)}\PY{p}{,} \PY{n}{axs}\PY{p}{[}\PY{n}{i}\PY{p}{]}\PY{o}{.}\PY{n}{axis}\PY{p}{(}\PY{l+s+s2}{\PYZdq{}}\PY{l+s+s2}{off}\PY{l+s+s2}{\PYZdq{}}\PY{p}{)}

    \PY{n}{plt}\PY{o}{.}\PY{n}{subplots\PYZus{}adjust}\PY{p}{(}\PY{n}{wspace}\PY{o}{=}\PY{l+m+mi}{0}\PY{p}{,} \PY{n}{hspace}\PY{o}{=}\PY{l+m+mi}{0}\PY{p}{)}
    \PY{n}{plt}\PY{o}{.}\PY{n}{show}\PY{p}{(}\PY{p}{)}

\PY{n}{visualize\PYZus{}list\PYZus{}of\PYZus{}samples}\PY{p}{(}\PY{p}{(}\PY{n}{flair\PYZus{}sample}\PY{p}{,} \PY{n}{gt\PYZus{}sample}\PY{p}{)}\PY{p}{)}
\end{Verbatim}
\end{tcolorbox}

    \section{Organizing the data}\label{organizing-the-data}

Now that we have defined the Sample object, which can represent any
\(x\) or \(y\), we could potentially start constructing multiple sets of
data pairs, e.g.~\(S'\), \(S''\), etc. We could do so by simply
having a list of lists of Sample objects. If we only had one data pair,
this would look as follows:

    \begin{tcolorbox}[breakable, size=fbox, boxrule=1pt, pad at break*=1mm,colback=cellbackground, colframe=cellborder]
\prompt{In}{incolor}{ }{\boxspacing}
\begin{Verbatim}[commandchars=\\\{\}]
\PY{n}{simple\PYZus{}set} \PY{o}{=} \PY{p}{[}\PY{p}{(}\PY{n}{flair\PYZus{}sample}\PY{p}{,} \PY{n}{gt\PYZus{}sample}\PY{p}{)}\PY{p}{]}
\end{Verbatim}
\end{tcolorbox}

    However, such a representation is neither valid nor feasible for various
reasons. Remember that the set \(\mathcal{S}'\) represents pairs of
native data, e.g.~samples drawn from the distribution
$P(X,Y)$. It would
become impossible to load everything into memory at once for large datasets or large data. Furthermore,
the subsequently derived sets \(\mathcal{S''}\), etc., are in fact
(practically) infinite, e.g.~data augmentation, and thus impossible to
represent with a simple list. We, therefore, chose to organize the
native data only, i.e.~the set \(\mathcal{S}'\), in a more structured manner.
The other sets are represented implicitly by defining the sampling
functions \(s''\), etc (see Section~\ref{data-sampling} and~\ref{data-transformation}).

To organize our data, we have created a hierarchical structure of
dictionaries: Mirc \textgreater{} Dataset \textgreater{} Case
\textgreater{} Record \textgreater{} Modality. When creating
instantiations of a Dataset, Case, Record or Modality, we need to
specify an ``ID''. As a result, each dictionary consists of key-value pairs,
with the keys corresponding to the lower-level object's IDs. The purpose
of the Mirc object is to group multiple Dataset objects and provide the
user with higher-level functionalities, of which we will explain a few at
the end of this section.

    \subsection{The Modality object}\label{the-modality-object}

Starting from the lowest level, a Modality is simply something that
returns a Sample object when \texttt{load()} is called. This is useful
because when having a large dataset, it is impossible to load every
image simultaneously; thus, the \texttt {} method provides a way
to only create a Sample object from it when needed. Note that the
Modality is an abstract class, and to use the
feature above, we should e.g.~use the NiftiFileModality. When
the dataset is small, or for e.g.~\emph{metadata} such as age, it could
be more efficient or appropriate to load everything to disk and use the
NiftiModality or ArrayModality instead. Other Modality subtypes that are
readily available are ImageFileModality and ImageFileMultiModality.
These use the Image library to load 2D images from disk (e.g.~PNG, JPG)
and convert them to 5D correctly (setting the fourth dimension as the
singleton dimension as mentioned before).

    \begin{tcolorbox}[breakable, size=fbox, boxrule=1pt, pad at break*=1mm,colback=cellbackground, colframe=cellborder]
\prompt{In}{incolor}{ }{\boxspacing}
\begin{Verbatim}[commandchars=\\\{\}]
\PY{k+kn}{from} \PY{n+nn}{deepvoxnet2}\PY{n+nn}{.}\PY{n+nn}{components}\PY{n+nn}{.}\PY{n+nn}{mirc} \PY{k+kn}{import} \PY{n}{NiftiFileModality}\PY{p}{,} \PY{n}{NiftiModality}\PY{p}{,} \PY{n}{ArrayModality}

\PY{n}{flair\PYZus{}modality} \PY{o}{=} \PY{n}{NiftiFileModality}\PY{p}{(}\PY{l+s+s2}{\PYZdq{}}\PY{l+s+s2}{flair}\PY{l+s+s2}{\PYZdq{}}\PY{p}{,} \PY{n}{file\PYZus{}path}\PY{o}{=}\PY{n}{flair\PYZus{}path}\PY{p}{)}
\PY{n}{flair\PYZus{}modality\PYZus{}} \PY{o}{=} \PY{n}{NiftiModality}\PY{p}{(}\PY{l+s+s2}{\PYZdq{}}\PY{l+s+s2}{flair}\PY{l+s+s2}{\PYZdq{}}\PY{p}{,} \PY{n}{nifty}\PY{o}{=}\PY{n}{flair\PYZus{}image}\PY{p}{)}
\PY{n}{flair\PYZus{}modality\PYZus{}\PYZus{}} \PY{o}{=} \PY{n}{ArrayModality}\PY{p}{(}\PY{l+s+s2}{\PYZdq{}}\PY{l+s+s2}{flair}\PY{l+s+s2}{\PYZdq{}}\PY{p}{,} \PY{n}{array}\PY{o}{=}\PY{n}{flair\PYZus{}image}\PY{o}{.}\PY{n}{get\PYZus{}fdata}\PY{p}{(}\PY{p}{)}\PY{p}{,} \PY{n}{affine}\PY{o}{=}\PY{n}{flair\PYZus{}image}\PY{o}{.}\PY{n}{affine}\PY{p}{)}
\PY{k}{assert} \PY{n}{np}\PY{o}{.}\PY{n}{array\PYZus{}equal}\PY{p}{(}\PY{n}{flair\PYZus{}modality}\PY{o}{.}\PY{n}{load}\PY{p}{(}\PY{p}{)}\PY{p}{,} \PY{n}{flair\PYZus{}modality\PYZus{}}\PY{o}{.}\PY{n}{load}\PY{p}{(}\PY{p}{)}\PY{p}{)}
\PY{k}{assert} \PY{n}{np}\PY{o}{.}\PY{n}{array\PYZus{}equal}\PY{p}{(}\PY{n}{flair\PYZus{}modality}\PY{o}{.}\PY{n}{load}\PY{p}{(}\PY{p}{)}\PY{p}{,} \PY{n}{flair\PYZus{}modality\PYZus{}\PYZus{}}\PY{o}{.}\PY{n}{load}\PY{p}{(}\PY{p}{)}\PY{p}{)}
\PY{n+nb}{print}\PY{p}{(}\PY{n}{flair\PYZus{}modality}\PY{o}{.}\PY{n}{load}\PY{p}{(}\PY{p}{)}\PY{o}{.}\PY{n}{shape}\PY{p}{)}
\PY{n+nb}{print}\PY{p}{(}\PY{n}{flair\PYZus{}modality}\PY{o}{.}\PY{n}{load}\PY{p}{(}\PY{p}{)}\PY{o}{.}\PY{n}{affine}\PY{p}{)}
\PY{n}{gt\PYZus{}modality} \PY{o}{=} \PY{n}{NiftiFileModality}\PY{p}{(}\PY{l+s+s2}{\PYZdq{}}\PY{l+s+s2}{gt}\PY{l+s+s2}{\PYZdq{}}\PY{p}{,} \PY{n}{file\PYZus{}path}\PY{o}{=}\PY{n}{flair\PYZus{}path}\PY{p}{)}
\PY{n}{metadata\PYZus{}modality} \PY{o}{=} \PY{n}{ArrayModality}\PY{p}{(}\PY{l+s+s2}{\PYZdq{}}\PY{l+s+s2}{age}\PY{l+s+s2}{\PYZdq{}}\PY{p}{,} \PY{n}{array}\PY{o}{=}\PY{l+m+mi}{45}\PY{p}{)}  \PY{c+c1}{\PYZsh{} just an example how we could make a metadata Sample, e.g. of age}
\PY{n+nb}{print}\PY{p}{(}\PY{n}{metadata\PYZus{}modality}\PY{o}{.}\PY{n}{load}\PY{p}{(}\PY{p}{)}\PY{o}{.}\PY{n}{shape}\PY{p}{)}
\PY{n+nb}{print}\PY{p}{(}\PY{n}{metadata\PYZus{}modality}\PY{o}{.}\PY{n}{load}\PY{p}{(}\PY{p}{)}\PY{o}{.}\PY{n}{affine}\PY{p}{)}
\end{Verbatim}
\end{tcolorbox}

    Notice that we have created a ``flair'' Modality in three ways but that
the final Sample objects are, in fact, the same. As such, always think of what is
the most efficient way to represent a particular data type and Sample
object in your specific situation.

    \subsection{The Record, Case and Dataset
objects}\label{the-record-case-and-dataset-objects}

To create ``pairs'' of data, we should group multiple Modality objects
under a so-called Record. Multiple Record objects need to
be grouped under a Case object. The idea of the existence of both a
Case and a Record is that there might be multiple \emph{records}
(e.g.~observations) for a certain \emph{case} (e.g.~a subject). Imagine
having multiple experts annotating the same subject or having multiple parts of the body scanned separately. Finally, cases
should be grouped under a Dataset object. Let us construct two tumor
datasets based on data from BRATS 2018 below.

    \begin{tcolorbox}[breakable, size=fbox, boxrule=1pt, pad at break*=1mm,colback=cellbackground, colframe=cellborder]
\prompt{In}{incolor}{ }{\boxspacing}
\begin{Verbatim}[commandchars=\\\{\}]
\PY{k+kn}{from} \PY{n+nn}{deepvoxnet2}\PY{n+nn}{.}\PY{n+nn}{components}\PY{n+nn}{.}\PY{n+nn}{mirc} \PY{k+kn}{import} \PY{n}{Dataset}\PY{p}{,} \PY{n}{Case}\PY{p}{,} \PY{n}{Record}

\PY{k}{def} \PY{n+nf}{get\PYZus{}tumor\PYZus{}dataset}\PY{p}{(}\PY{n}{dataset\PYZus{}id}\PY{p}{,} \PY{n}{case\PYZus{}indices}\PY{p}{)}\PY{p}{:}
    \PY{n}{tumor\PYZus{}dataset} \PY{o}{=} \PY{n}{Dataset}\PY{p}{(}\PY{n}{dataset\PYZus{}id}\PY{p}{)}
    \PY{k}{for} \PY{n}{i} \PY{o+ow}{in} \PY{n}{case\PYZus{}indices}\PY{p}{:}
        \PY{n}{subject} \PY{o}{=} \PY{n}{Case}\PY{p}{(}\PY{l+s+sa}{f}\PY{l+s+s2}{\PYZdq{}}\PY{l+s+s2}{subject\PYZus{}}\PY{l+s+si}{\PYZob{}}\PY{n}{i}\PY{l+s+si}{\PYZcb{}}\PY{l+s+s2}{\PYZdq{}}\PY{p}{)}
        \PY{k}{for} \PY{n}{j} \PY{o+ow}{in} \PY{n+nb}{range}\PY{p}{(}\PY{l+m+mi}{1}\PY{p}{)}\PY{p}{:}  \PY{c+c1}{\PYZsh{} we only have a single observation for each subject here}
            \PY{n}{observation} \PY{o}{=} \PY{n}{Record}\PY{p}{(}\PY{l+s+sa}{f}\PY{l+s+s2}{\PYZdq{}}\PY{l+s+s2}{observation\PYZus{}}\PY{l+s+si}{\PYZob{}}\PY{n}{j}\PY{l+s+si}{\PYZcb{}}\PY{l+s+s2}{\PYZdq{}}\PY{p}{)}
            \PY{n}{flair\PYZus{}modality} \PY{o}{=} \PY{n}{NiftiFileModality}\PY{p}{(}\PY{l+s+s2}{\PYZdq{}}\PY{l+s+s2}{flair}\PY{l+s+s2}{\PYZdq{}}\PY{p}{,} \PY{n}{file\PYZus{}path}\PY{o}{=}\PY{n}{os}\PY{o}{.}\PY{n}{path}\PY{o}{.}\PY{n}{join}\PY{p}{(}\PY{n}{DEMO\PYZus{}DIR}\PY{p}{,} \PY{l+s+s2}{\PYZdq{}}\PY{l+s+s2}{brats\PYZus{}2018}\PY{l+s+s2}{\PYZdq{}}\PY{p}{,} \PY{l+s+sa}{f}\PY{l+s+s2}{\PYZdq{}}\PY{l+s+s2}{case\PYZus{}}\PY{l+s+si}{\PYZob{}}\PY{n}{i}\PY{l+s+si}{\PYZcb{}}\PY{l+s+s2}{\PYZdq{}}\PY{p}{,} \PY{l+s+s2}{\PYZdq{}}\PY{l+s+s2}{FLAIR.nii.gz}\PY{l+s+s2}{\PYZdq{}}\PY{p}{)}\PY{p}{)}
            \PY{n}{gt\PYZus{}modality} \PY{o}{=} \PY{n}{NiftiFileModality}\PY{p}{(}\PY{l+s+s2}{\PYZdq{}}\PY{l+s+s2}{gt}\PY{l+s+s2}{\PYZdq{}}\PY{p}{,} \PY{n}{file\PYZus{}path}\PY{o}{=}\PY{n}{os}\PY{o}{.}\PY{n}{path}\PY{o}{.}\PY{n}{join}\PY{p}{(}\PY{n}{DEMO\PYZus{}DIR}\PY{p}{,} \PY{l+s+s2}{\PYZdq{}}\PY{l+s+s2}{brats\PYZus{}2018}\PY{l+s+s2}{\PYZdq{}}\PY{p}{,} \PY{l+s+sa}{f}\PY{l+s+s2}{\PYZdq{}}\PY{l+s+s2}{case\PYZus{}}\PY{l+s+si}{\PYZob{}}\PY{n}{i}\PY{l+s+si}{\PYZcb{}}\PY{l+s+s2}{\PYZdq{}}\PY{p}{,} \PY{l+s+s2}{\PYZdq{}}\PY{l+s+s2}{GT\PYZus{}W.nii.gz}\PY{l+s+s2}{\PYZdq{}}\PY{p}{)}\PY{p}{)}
            \PY{n}{age\PYZus{}modality} \PY{o}{=} \PY{n}{ArrayModality}\PY{p}{(}\PY{l+s+s2}{\PYZdq{}}\PY{l+s+s2}{age}\PY{l+s+s2}{\PYZdq{}}\PY{p}{,} \PY{n}{np}\PY{o}{.}\PY{n}{random}\PY{o}{.}\PY{n}{randint}\PY{p}{(}\PY{l+m+mi}{20}\PY{p}{,} \PY{l+m+mi}{80}\PY{p}{)}\PY{p}{)}
            \PY{n}{observation}\PY{o}{.}\PY{n}{add}\PY{p}{(}\PY{n}{flair\PYZus{}modality}\PY{p}{)}
            \PY{n}{observation}\PY{o}{.}\PY{n}{add}\PY{p}{(}\PY{n}{gt\PYZus{}modality}\PY{p}{)}
            \PY{n}{observation}\PY{o}{.}\PY{n}{add}\PY{p}{(}\PY{n}{age\PYZus{}modality}\PY{p}{)}
            \PY{n}{subject}\PY{o}{.}\PY{n}{add}\PY{p}{(}\PY{n}{observation}\PY{p}{)}

        \PY{n}{tumor\PYZus{}dataset}\PY{o}{.}\PY{n}{add}\PY{p}{(}\PY{n}{subject}\PY{p}{)}

    \PY{k}{return} \PY{n}{tumor\PYZus{}dataset}

\PY{n}{train\PYZus{}tumor\PYZus{}dataset} \PY{o}{=} \PY{n}{get\PYZus{}tumor\PYZus{}dataset}\PY{p}{(}\PY{l+s+s2}{\PYZdq{}}\PY{l+s+s2}{train\PYZus{}dataset}\PY{l+s+s2}{\PYZdq{}}\PY{p}{,} \PY{n+nb}{range}\PY{p}{(}\PY{l+m+mi}{0}\PY{p}{,} \PY{l+m+mi}{8}\PY{p}{)}\PY{p}{)}
\PY{n}{val\PYZus{}tumor\PYZus{}dataset} \PY{o}{=} \PY{n}{get\PYZus{}tumor\PYZus{}dataset}\PY{p}{(}\PY{l+s+s2}{\PYZdq{}}\PY{l+s+s2}{val\PYZus{}dataset}\PY{l+s+s2}{\PYZdq{}}\PY{p}{,} \PY{n+nb}{range}\PY{p}{(}\PY{l+m+mi}{8}\PY{p}{,} \PY{l+m+mi}{10}\PY{p}{)}\PY{p}{)}
\PY{k}{for} \PY{n}{tumor\PYZus{}dataset} \PY{o+ow}{in} \PY{p}{[}\PY{n}{train\PYZus{}tumor\PYZus{}dataset}\PY{p}{,} \PY{n}{val\PYZus{}tumor\PYZus{}dataset}\PY{p}{]}\PY{p}{:}
    \PY{n+nb}{print}\PY{p}{(}\PY{l+s+sa}{f}\PY{l+s+s2}{\PYZdq{}}\PY{l+s+si}{\PYZob{}}\PY{n}{tumor\PYZus{}dataset}\PY{o}{.}\PY{n}{dataset\PYZus{}id}\PY{l+s+si}{\PYZcb{}}\PY{l+s+s2}{ contains }\PY{l+s+si}{\PYZob{}}\PY{n+nb}{len}\PY{p}{(}\PY{n}{tumor\PYZus{}dataset}\PY{p}{)}\PY{l+s+si}{\PYZcb{}}\PY{l+s+s2}{ cases:}\PY{l+s+s2}{\PYZdq{}}\PY{p}{)}
    \PY{k}{for} \PY{n}{case\PYZus{}id} \PY{o+ow}{in} \PY{n}{tumor\PYZus{}dataset}\PY{p}{:}
        \PY{n+nb}{print}\PY{p}{(}\PY{l+s+sa}{f}\PY{l+s+s2}{\PYZdq{}}\PY{l+s+se}{\PYZbs{}t}\PY{l+s+si}{\PYZob{}}\PY{n}{case\PYZus{}id}\PY{l+s+si}{\PYZcb{}}\PY{l+s+s2}{ contains }\PY{l+s+si}{\PYZob{}}\PY{n+nb}{len}\PY{p}{(}\PY{n}{tumor\PYZus{}dataset}\PY{p}{[}\PY{n}{case\PYZus{}id}\PY{p}{]}\PY{p}{)}\PY{l+s+si}{\PYZcb{}}\PY{l+s+s2}{ records:}\PY{l+s+s2}{\PYZdq{}}\PY{p}{)}
        \PY{k}{for} \PY{n}{record\PYZus{}id} \PY{o+ow}{in} \PY{n}{tumor\PYZus{}dataset}\PY{p}{[}\PY{n}{case\PYZus{}id}\PY{p}{]}\PY{p}{:}
            \PY{n+nb}{print}\PY{p}{(}\PY{l+s+sa}{f}\PY{l+s+s2}{\PYZdq{}}\PY{l+s+se}{\PYZbs{}t}\PY{l+s+se}{\PYZbs{}t}\PY{l+s+si}{\PYZob{}}\PY{n}{record\PYZus{}id}\PY{l+s+si}{\PYZcb{}}\PY{l+s+s2}{ contains }\PY{l+s+si}{\PYZob{}}\PY{n+nb}{len}\PY{p}{(}\PY{n}{tumor\PYZus{}dataset}\PY{p}{[}\PY{n}{case\PYZus{}id}\PY{p}{]}\PY{p}{[}\PY{n}{record\PYZus{}id}\PY{p}{]}\PY{p}{)}\PY{l+s+si}{\PYZcb{}}\PY{l+s+s2}{ modalities:}\PY{l+s+s2}{\PYZdq{}}\PY{p}{)}
            \PY{k}{for} \PY{n}{modality\PYZus{}id} \PY{o+ow}{in} \PY{n}{tumor\PYZus{}dataset}\PY{p}{[}\PY{n}{case\PYZus{}id}\PY{p}{]}\PY{p}{[}\PY{n}{record\PYZus{}id}\PY{p}{]}\PY{p}{:}
                \PY{n+nb}{print}\PY{p}{(}\PY{l+s+sa}{f}\PY{l+s+s2}{\PYZdq{}}\PY{l+s+se}{\PYZbs{}t}\PY{l+s+se}{\PYZbs{}t}\PY{l+s+se}{\PYZbs{}t}\PY{l+s+si}{\PYZob{}}\PY{n}{modality\PYZus{}id}\PY{l+s+si}{\PYZcb{}}\PY{l+s+s2}{\PYZdq{}}\PY{p}{)}
\end{Verbatim}
\end{tcolorbox}

    \subsection{The Mirc object}\label{the-mirc-object}

When we have built our dataset, we can group it, optionally with other
datasets, into a Mirc object. The Mirc object is a powerful tool for getting
a complete overview of all your data. Calling
\texttt{get\_dataset\_ids()}, \texttt{get\_case\_ids()},
\texttt{get\_record\_ids()} or \texttt{get\_modality\_ids()} you can get
a quick overview on what data is available in your Mirc object. You can
also calculate the mean and standard deviation for a particular modality
by using the \texttt{mean\_and\_std} method. The \texttt {inspect} method
is particularly useful to check multiple modalities for spatial
consistency within a certain Record and get an overview of the
distribution of voxel sizes and other spatial characteristics of your
data. Using the \texttt{get\_df} method, it is possible to acquire a
Pandas DataFrame such that you could do other types of analyses or
inspections. This is particularly useful if you want to inspect
distributions of non-imaging data, such as age.

    \begin{tcolorbox}[breakable, size=fbox, boxrule=1pt, pad at break*=1mm,colback=cellbackground, colframe=cellborder]
\prompt{In}{incolor}{ }{\boxspacing}
\begin{Verbatim}[commandchars=\\\{\}]
\PY{k+kn}{from} \PY{n+nn}{deepvoxnet2}\PY{n+nn}{.}\PY{n+nn}{components}\PY{n+nn}{.}\PY{n+nn}{mirc} \PY{k+kn}{import} \PY{n}{Mirc}

\PY{n}{my\PYZus{}mirc} \PY{o}{=} \PY{n}{Mirc}\PY{p}{(}\PY{n}{train\PYZus{}tumor\PYZus{}dataset}\PY{p}{,} \PY{n}{val\PYZus{}tumor\PYZus{}dataset}\PY{p}{)}
\PY{n+nb}{print}\PY{p}{(}\PY{n}{my\PYZus{}mirc}\PY{o}{.}\PY{n}{get\PYZus{}dataset\PYZus{}ids}\PY{p}{(}\PY{p}{)}\PY{p}{)}
\PY{n+nb}{print}\PY{p}{(}\PY{n}{my\PYZus{}mirc}\PY{o}{.}\PY{n}{get\PYZus{}case\PYZus{}ids}\PY{p}{(}\PY{p}{)}\PY{p}{)}
\PY{n+nb}{print}\PY{p}{(}\PY{n}{my\PYZus{}mirc}\PY{o}{.}\PY{n}{get\PYZus{}record\PYZus{}ids}\PY{p}{(}\PY{p}{)}\PY{p}{)}
\PY{n+nb}{print}\PY{p}{(}\PY{n}{my\PYZus{}mirc}\PY{o}{.}\PY{n}{get\PYZus{}modality\PYZus{}ids}\PY{p}{(}\PY{p}{)}\PY{p}{)}
\PY{n+nb}{print}\PY{p}{(}\PY{n}{my\PYZus{}mirc}\PY{o}{.}\PY{n}{mean\PYZus{}and\PYZus{}std}\PY{p}{(}\PY{l+s+s2}{\PYZdq{}}\PY{l+s+s2}{flair}\PY{l+s+s2}{\PYZdq{}}\PY{p}{,} \PY{n}{n}\PY{o}{=}\PY{l+m+mi}{2}\PY{p}{)}\PY{p}{)}  \PY{c+c1}{\PYZsh{} computing the mean and standard deviation using all your images might be too expensive, therefore limit the number of images used setting the n option}
\PY{n+nb}{print}\PY{p}{(}\PY{n}{my\PYZus{}mirc}\PY{o}{.}\PY{n}{inspect}\PY{p}{(}\PY{p}{[}\PY{l+s+s2}{\PYZdq{}}\PY{l+s+s2}{flair}\PY{l+s+s2}{\PYZdq{}}\PY{p}{,} \PY{l+s+s2}{\PYZdq{}}\PY{l+s+s2}{gt}\PY{l+s+s2}{\PYZdq{}}\PY{p}{]}\PY{p}{,} \PY{n}{ns}\PY{o}{=}\PY{p}{[}\PY{l+m+mi}{2}\PY{p}{,} \PY{l+m+mi}{2}\PY{p}{]}\PY{p}{)}\PY{p}{)}
\PY{n+nb}{print}\PY{p}{(}\PY{n}{my\PYZus{}mirc}\PY{o}{.}\PY{n}{get\PYZus{}df}\PY{p}{(}\PY{l+s+s2}{\PYZdq{}}\PY{l+s+s2}{age}\PY{l+s+s2}{\PYZdq{}}\PY{p}{)}\PY{p}{)}
\end{Verbatim}
\end{tcolorbox}

    In the usage of the Mirc object example mentioned above, we have grouped
both our training and our validation set under one Mirc object. Keep in
mind that this was just for illustration purposes. It is always a good
idea to keep the two separate.

    \begin{tcolorbox}[breakable, size=fbox, boxrule=1pt, pad at break*=1mm,colback=cellbackground, colframe=cellborder]
\prompt{In}{incolor}{ }{\boxspacing}
\begin{Verbatim}[commandchars=\\\{\}]
\PY{n}{my\PYZus{}train\PYZus{}mirc} \PY{o}{=} \PY{n}{Mirc}\PY{p}{(}\PY{n}{train\PYZus{}tumor\PYZus{}dataset}\PY{p}{)}
\PY{n}{my\PYZus{}val\PYZus{}mirc} \PY{o}{=} \PY{n}{Mirc}\PY{p}{(}\PY{n}{val\PYZus{}tumor\PYZus{}dataset}\PY{p}{)}
\end{Verbatim}
\end{tcolorbox}

    \section{Data sampling}\label{data-sampling}

Once we have defined the set \(\mathcal{S'}\), e.g.~by means of a
Mirc object containing Record objects at its lowest level or simply a
list of lists of Sample objects, we can start thinking about the
sampling functions, e.g.~the function \(s''\) that produces the set
\(\mathcal{S''}\) from \(\mathcal{S'}\) (Equation~\ref{eq:dvn_eq}). It turns out that we can often
view any sampling function that goes from one set of data pairs to
another as something that first samples a data pair and then applies
some form of data transformation to it creating zero, one or more
transformed data pairs. In this section, we will discuss the first part
of the sampling function, more specifically, the Sampler object. For now,
you can view the Sampler object as having an internal list of objects
that we can index to retrieve the object. While the retrieved object
could be a Record object or a list of Sample objects, we chose to define
it as a new object, i.e.~the Identifier object.

\subsection{The Identifier object}\label{the-identifier-object}

Let us start by saying that the Identifier object could be any
object since we could define the word \emph{sampling} as retrieving
\emph{something} from a list. In our specific use case, the returned
Identifier object will be used by the data transformation function (see
Section~\ref{data-transformation}), and thus should contain all information necessary to
do the transformation. Letting the Identifier represent a list of Samples
or a Record would be sufficient in that case. However, if representing a
Record, the Identifier has no notion from which Dataset or Case it
was from. For this purpose, if we want to keep track of this additional
information, we could wrap a Record as a MircIdentifier.

    \begin{tcolorbox}[breakable, size=fbox, boxrule=1pt, pad at break*=1mm,colback=cellbackground, colframe=cellborder]
\prompt{In}{incolor}{ }{\boxspacing}
\begin{Verbatim}[commandchars=\\\{\}]
\PY{k+kn}{from} \PY{n+nn}{deepvoxnet2}\PY{n+nn}{.}\PY{n+nn}{components}\PY{n+nn}{.}\PY{n+nn}{sampler} \PY{k+kn}{import} \PY{n}{MircIdentifier}

\PY{n}{record\PYZus{}identifier} \PY{o}{=} \PY{n}{MircIdentifier}\PY{p}{(}\PY{n}{my\PYZus{}mirc}\PY{p}{,} \PY{n}{dataset\PYZus{}id}\PY{o}{=}\PY{l+s+s2}{\PYZdq{}}\PY{l+s+s2}{train\PYZus{}dataset}\PY{l+s+s2}{\PYZdq{}}\PY{p}{,} \PY{n}{case\PYZus{}id}\PY{o}{=}\PY{l+s+s2}{\PYZdq{}}\PY{l+s+s2}{subject\PYZus{}0}\PY{l+s+s2}{\PYZdq{}}\PY{p}{,} \PY{n}{record\PYZus{}id}\PY{o}{=}\PY{l+s+s2}{\PYZdq{}}\PY{l+s+s2}{observation\PYZus{}0}\PY{l+s+s2}{\PYZdq{}}\PY{p}{)}
\PY{n+nb}{print}\PY{p}{(}\PY{n}{record\PYZus{}identifier}\PY{p}{(}\PY{p}{)}\PY{p}{)}
\PY{k}{for} \PY{n}{modality\PYZus{}id} \PY{o+ow}{in} \PY{n}{record\PYZus{}identifier}\PY{o}{.}\PY{n}{mirc}\PY{p}{[}\PY{n}{record\PYZus{}identifier}\PY{o}{.}\PY{n}{dataset\PYZus{}id}\PY{p}{]}\PY{p}{[}\PY{n}{record\PYZus{}identifier}\PY{o}{.}\PY{n}{case\PYZus{}id}\PY{p}{]}\PY{p}{[}\PY{n}{record\PYZus{}identifier}\PY{o}{.}\PY{n}{record\PYZus{}id}\PY{p}{]}\PY{p}{:}
    \PY{n+nb}{print}\PY{p}{(}\PY{n}{modality\PYZus{}id}\PY{p}{)}
\end{Verbatim}
\end{tcolorbox}

    \subsection{The Sampler object}\label{the-sampler-object}

As mentioned before, we defined a Sampler as a rather simple
object, specifically a list of Identifier objects, but with some
additional functionalities. An important functionality is the ability to
\emph{shuffle} its internal list of Identifier objects by calling
\texttt{randomize()}. Whether this call will do anything depends on the
value of the \texttt{shuffle} option when you create the Sampler, which
defaults to \texttt{False}. The shuffling itself is a new permutation of
the list of Identifiers, potentially with replacement according to the
probability density if specified using the \texttt{weights} option. In
\gls{dvn2} you can use the MircSampler to convert a Mirc object to a Sampler
of MircIdentifiers. Depending on the \texttt{mode} option, either
\texttt{"per\_case"} or \texttt{"per\_record"}, the internal list of
MircIdentifiers and its shuffling will be different. Using
\texttt{mode="per\_case"}, the list size will be the number of
Case objects in the Mirc object, thus with Record objects selected
uniformly across all Case objects. Using \texttt{mode="per\_record"}, the
list size will be the number of Record objects in the Mirc
object, thus with Records selected uniformly across all Record objects.

    \begin{tcolorbox}[breakable, size=fbox, boxrule=1pt, pad at break*=1mm,colback=cellbackground, colframe=cellborder]
\prompt{In}{incolor}{ }{\boxspacing}
\begin{Verbatim}[commandchars=\\\{\}]
\PY{k+kn}{from} \PY{n+nn}{deepvoxnet2}\PY{n+nn}{.}\PY{n+nn}{components}\PY{n+nn}{.}\PY{n+nn}{sampler} \PY{k+kn}{import} \PY{n}{Sampler}\PY{p}{,} \PY{n}{MircSampler}

\PY{n}{a\PYZus{}simple\PYZus{}sampler} \PY{o}{=} \PY{n}{Sampler}\PY{p}{(}\PY{p}{[}\PY{l+m+mi}{1}\PY{p}{,} \PY{l+m+mi}{2}\PY{p}{,} \PY{l+m+mi}{3}\PY{p}{,} \PY{l+m+mi}{4}\PY{p}{,} \PY{l+m+mi}{5}\PY{p}{]}\PY{p}{,} \PY{n}{shuffle}\PY{o}{=}\PY{k+kc}{True}\PY{p}{,} \PY{n}{weights}\PY{o}{=}\PY{p}{[}\PY{l+m+mi}{1}\PY{p}{,} \PY{l+m+mi}{1}\PY{p}{,} \PY{l+m+mi}{10}\PY{p}{,} \PY{l+m+mi}{1}\PY{p}{,} \PY{l+m+mi}{1}\PY{p}{]}\PY{p}{)}
\PY{k}{for} \PY{n}{i} \PY{o+ow}{in} \PY{n+nb}{range}\PY{p}{(}\PY{l+m+mi}{5}\PY{p}{)}\PY{p}{:}
    \PY{n+nb}{print}\PY{p}{(}\PY{l+s+sa}{f}\PY{l+s+s2}{\PYZdq{}}\PY{l+s+s2}{a\PYZus{}simple\PYZus{}sampler at iteration }\PY{l+s+si}{\PYZob{}}\PY{n}{i}\PY{l+s+si}{\PYZcb{}}\PY{l+s+s2}{: }\PY{l+s+s2}{\PYZdq{}}\PY{p}{,} \PY{p}{[}\PY{n}{a\PYZus{}simple\PYZus{}sampler}\PY{p}{[}\PY{n}{j}\PY{p}{]} \PY{k}{for} \PY{n}{j} \PY{o+ow}{in} \PY{n+nb}{range}\PY{p}{(}\PY{n+nb}{len}\PY{p}{(}\PY{n}{a\PYZus{}simple\PYZus{}sampler}\PY{p}{)}\PY{p}{)}\PY{p}{]}\PY{p}{)}
    \PY{n}{a\PYZus{}simple\PYZus{}sampler}\PY{o}{.}\PY{n}{randomize}\PY{p}{(}\PY{p}{)}

\PY{n}{my\PYZus{}train\PYZus{}sampler} \PY{o}{=} \PY{n}{MircSampler}\PY{p}{(}\PY{n}{my\PYZus{}train\PYZus{}mirc}\PY{p}{,} \PY{n}{mode}\PY{o}{=}\PY{l+s+s2}{\PYZdq{}}\PY{l+s+s2}{per\PYZus{}case}\PY{l+s+s2}{\PYZdq{}}\PY{p}{,} \PY{n}{shuffle}\PY{o}{=}\PY{k+kc}{True}\PY{p}{)}  \PY{c+c1}{\PYZsh{} shuffle=True typically for training}
\PY{n}{my\PYZus{}val\PYZus{}sampler} \PY{o}{=} \PY{n}{MircSampler}\PY{p}{(}\PY{n}{my\PYZus{}val\PYZus{}mirc}\PY{p}{,} \PY{n}{mode}\PY{o}{=}\PY{l+s+s2}{\PYZdq{}}\PY{l+s+s2}{per\PYZus{}case}\PY{l+s+s2}{\PYZdq{}}\PY{p}{,} \PY{n}{shuffle}\PY{o}{=}\PY{k+kc}{False}\PY{p}{)}  \PY{c+c1}{\PYZsh{} shuffle=False typically for validation}
\PY{k}{for} \PY{n}{i} \PY{o+ow}{in} \PY{n+nb}{range}\PY{p}{(}\PY{l+m+mi}{5}\PY{p}{)}\PY{p}{:}
    \PY{n+nb}{print}\PY{p}{(}\PY{l+s+sa}{f}\PY{l+s+s2}{\PYZdq{}}\PY{l+s+s2}{my\PYZus{}train\PYZus{}sampler at iteration }\PY{l+s+si}{\PYZob{}}\PY{n}{i}\PY{l+s+si}{\PYZcb{}}\PY{l+s+s2}{: }\PY{l+s+s2}{\PYZdq{}}\PY{p}{,} \PY{p}{[}\PY{n}{my\PYZus{}train\PYZus{}sampler}\PY{p}{[}\PY{n}{j}\PY{p}{]}\PY{o}{.}\PY{n}{case\PYZus{}id} \PY{k}{for} \PY{n}{j} \PY{o+ow}{in} \PY{n+nb}{range}\PY{p}{(}\PY{n+nb}{len}\PY{p}{(}\PY{n}{my\PYZus{}train\PYZus{}sampler}\PY{p}{)}\PY{p}{)}\PY{p}{]}\PY{p}{)}
    \PY{n+nb}{print}\PY{p}{(}\PY{l+s+sa}{f}\PY{l+s+s2}{\PYZdq{}}\PY{l+s+s2}{my\PYZus{}val\PYZus{}sampler at iteration }\PY{l+s+si}{\PYZob{}}\PY{n}{i}\PY{l+s+si}{\PYZcb{}}\PY{l+s+s2}{: }\PY{l+s+s2}{\PYZdq{}}\PY{p}{,} \PY{p}{[}\PY{n}{my\PYZus{}val\PYZus{}sampler}\PY{p}{[}\PY{n}{j}\PY{p}{]}\PY{o}{.}\PY{n}{case\PYZus{}id} \PY{k}{for} \PY{n}{j} \PY{o+ow}{in} \PY{n+nb}{range}\PY{p}{(}\PY{n+nb}{len}\PY{p}{(}\PY{n}{my\PYZus{}val\PYZus{}sampler}\PY{p}{)}\PY{p}{)}\PY{p}{]}\PY{p}{)}
    \PY{n}{my\PYZus{}train\PYZus{}sampler}\PY{o}{.}\PY{n}{randomize}\PY{p}{(}\PY{p}{)}
    \PY{n}{my\PYZus{}val\PYZus{}sampler}\PY{o}{.}\PY{n}{randomize}\PY{p}{(}\PY{p}{)}
\end{Verbatim}
\end{tcolorbox}

    It might look strange that a Sampler has a finite internal list and is
not just something that \emph{generates} Identifier objects indefinitely
or until completion. However, using this abstraction, it is easier to
keep track of having seen all data pairs at least once before shuffling
(without \texttt{weights} specified), etc.

    \section{Data transformation}\label{data-transformation}

Now that we have selected a data pair from our set \(S'\), e.g.~as a
MircIdentifier using a MircSampler constructed from a Mirc object, we
want to transform this data pair into zero, one or more transformed data
pairs. To do so, \gls{dvn2} uses Transformer objects. To achieve complex
transformations, a network of Transformer objects can be constructed and
wrapped as a Creator object to unlock higher-level functionalities.

To start, we can view a Transformer object as having an input and an
output using the most straightforward representation of a data pair, i.e.~a
list of Sample objects. Furthermore, from its input, it can generate
zero, one or more (potentially infinite) outputs. What transformation
and how many strictly depends on the type of Transformer and what options
you specify. Typically, the input of one Transformer is the output of
another.

\subsection{The InputTransformer
object}\label{the-inputtransformer-object}

While all Transformers work as input-to-output and use a
list-of-Sample-objects representation, one type of Transformer
is slightly different: the InputTransformer. In a sense, the
InputTransformer has no input and thus \review{cannot} be connected to the
output of another Transformer. They serve as the start of your
Transformer network and exist to convert any type of Identifier into a
list of Sample objects. It is this property that makes it possible to
work with a more structured form of your dataset, using any type of
Identifier produced by the Sampler. The sole thing that must be
foreseen is a dedicated InputTransformer that knows how to convert the
Identifier to a list of Sample objects that it will present at its
output. When making use of MircIdentifiers, we can use the MircInput
InputTransformer. When using a list of Sample objects already,
we should use the SampleInput InputTransformer. Typically for any
InputTransformer is that we need to load the Identifier into the
Transformer \emph{manually} by using its \texttt{load} method.

    \begin{tcolorbox}[breakable, size=fbox, boxrule=1pt, pad at break*=1mm,colback=cellbackground, colframe=cellborder]
\prompt{In}{incolor}{ }{\boxspacing}
\begin{Verbatim}[commandchars=\\\{\}]
\PY{k+kn}{from} \PY{n+nn}{deepvoxnet2}\PY{n+nn}{.}\PY{n+nn}{components}\PY{n+nn}{.}\PY{n+nn}{transformers} \PY{k+kn}{import} \PY{n}{\PYZus{}MircInput}

\PY{n}{mirc\PYZus{}input} \PY{o}{=} \PY{n}{\PYZus{}MircInput}\PY{p}{(}\PY{p}{[}\PY{l+s+s2}{\PYZdq{}}\PY{l+s+s2}{flair}\PY{l+s+s2}{\PYZdq{}}\PY{p}{,} \PY{l+s+s2}{\PYZdq{}}\PY{l+s+s2}{gt}\PY{l+s+s2}{\PYZdq{}}\PY{p}{]}\PY{p}{,} \PY{n}{n}\PY{o}{=}\PY{l+m+mi}{1}\PY{p}{)}  \PY{c+c1}{\PYZsh{} the use of \PYZus{}MircInput(...) is for illustrative purposes only}
\PY{n}{mirc\PYZus{}input}\PY{o}{.}\PY{n}{load}\PY{p}{(}\PY{n}{my\PYZus{}val\PYZus{}sampler}\PY{p}{[}\PY{l+m+mi}{1}\PY{p}{]}\PY{p}{)}  \PY{c+c1}{\PYZsh{} here we do this manually, but usually we use the higher\PYZhy{}level Creator, which we can also use for debugging}
\PY{n}{x\PYZus{}y} \PY{o}{=} \PY{n}{mirc\PYZus{}input}\PY{p}{(}\PY{p}{)}  \PY{c+c1}{\PYZsh{} normally we will use MircInput(...) instead, which does \PYZus{}MircInput(...)() in one go}

\PY{k}{def} \PY{n+nf}{visualize\PYZus{}transformer\PYZus{}output}\PY{p}{(}\PY{n}{transformer\PYZus{}output}\PY{p}{)}\PY{p}{:}
    \PY{k}{for} \PY{n}{i} \PY{o+ow}{in} \PY{n+nb}{range}\PY{p}{(}\PY{l+m+mi}{100}\PY{p}{)}\PY{p}{:}
        \PY{k}{try}\PY{p}{:}
            \PY{n}{transformer\PYZus{}output\PYZus{}i} \PY{o}{=} \PY{n}{transformer\PYZus{}output}\PY{o}{.}\PY{n}{eval}\PY{p}{(}\PY{p}{)}
            \PY{n+nb}{print}\PY{p}{(}\PY{l+s+sa}{f}\PY{l+s+s2}{\PYZdq{}}\PY{l+s+s2}{Evaluation }\PY{l+s+si}{\PYZob{}}\PY{n}{i}\PY{l+s+si}{\PYZcb{}}\PY{l+s+s2}{:}\PY{l+s+s2}{\PYZdq{}}\PY{p}{)}
            \PY{n}{visualize\PYZus{}list\PYZus{}of\PYZus{}samples}\PY{p}{(}\PY{n}{transformer\PYZus{}output\PYZus{}i}\PY{p}{)}

        \PY{k}{except} \PY{n+ne}{StopIteration}\PY{p}{:}
            \PY{n+nb}{print}\PY{p}{(}\PY{l+s+sa}{f}\PY{l+s+s2}{\PYZdq{}}\PY{l+s+s2}{Generation stopped after n=}\PY{l+s+si}{\PYZob{}}\PY{n}{i}\PY{l+s+si}{\PYZcb{}}\PY{l+s+s2}{ evaluations}\PY{l+s+s2}{\PYZdq{}}\PY{p}{)}
            \PY{k}{break}

\PY{n}{visualize\PYZus{}transformer\PYZus{}output}\PY{p}{(}\PY{n}{x\PYZus{}y}\PY{p}{)}
\end{Verbatim}
\end{tcolorbox}

    In the above example, the InputTransformer ran out of generating new
data pairs after one iteration. In fact, by setting the \texttt{n}
option differently (\texttt{n=1} by default for most Transformers), you
can let the Transformer generate n outputs from one input.

    \subsection{The Creator object}\label{the-creator-object}

Suppose we not only want to load the Identifier but also want to
\emph{transform} it, such that in combination with the Sampler, we create
an implicit new set of data pairs. For this purpose, the returned output
of any Transformer, e.g.~\texttt{x\_y} in the above code snippet, is a
Connection object that can be \emph{connected} to the input of another
Transformer. Without going into too much detail, it is simply a reference
to a list of Sample objects at a certain output index (a Transformer can
have multiple inputs and outputs; see later) of a certain Transformer.
It is this abstraction that allows us to build a network of
Transformers. Suppose we want to create a new set in which we remove the
Y and only keep X, or vice versa:

    \begin{tcolorbox}[breakable, size=fbox, boxrule=1pt, pad at break*=1mm,colback=cellbackground, colframe=cellborder]
\prompt{In}{incolor}{ }{\boxspacing}
\begin{Verbatim}[commandchars=\\\{\}]
\PY{k+kn}{from} \PY{n+nn}{deepvoxnet2}\PY{n+nn}{.}\PY{n+nn}{components}\PY{n+nn}{.}\PY{n+nn}{transformers} \PY{k+kn}{import} \PY{n}{Split}

\PY{n}{x} \PY{o}{=} \PY{n}{Split}\PY{p}{(}\PY{n}{indices}\PY{o}{=}\PY{p}{(}\PY{l+m+mi}{0}\PY{p}{,}\PY{p}{)}\PY{p}{,} \PY{n}{n}\PY{o}{=}\PY{l+m+mi}{1}\PY{p}{)}\PY{p}{(}\PY{n}{x\PYZus{}y}\PY{p}{)}
\PY{n}{visualize\PYZus{}transformer\PYZus{}output}\PY{p}{(}\PY{n}{x}\PY{p}{)}
\PY{n}{y} \PY{o}{=} \PY{n}{Split}\PY{p}{(}\PY{n}{indices}\PY{o}{=}\PY{p}{(}\PY{l+m+mi}{1}\PY{p}{,}\PY{p}{)}\PY{p}{,} \PY{n}{n}\PY{o}{=}\PY{l+m+mi}{1}\PY{p}{)}\PY{p}{(}\PY{n}{x\PYZus{}y}\PY{p}{)}
\PY{n}{visualize\PYZus{}transformer\PYZus{}output}\PY{p}{(}\PY{n}{y}\PY{p}{)}
\end{Verbatim}
\end{tcolorbox}

    Even though we ask to produce a total of N=1*1=1 transformed data pairs
(for each new identifier), the generation stops immediately. This is
because we already ran that same MircInput object
(i.e.~\emph{mirc\_input}) until completion in the previous code snippet.
As soon as the Split Transformer wants to read the output of
\emph{mirc\_input}, the latter will tell it has already been depleted.
To overcome this burden, e.g.~when debugging or when the Transformer
network has run until completion for Identfier A, but we want to reuse it
and run it for Identifier B, we can use the Creator object. The Creator
object has a higher-level view of the Transformer network. For example,
it can find all InputTransformer objects in the Transformer network and
can reset the state of all Transformer objects. Below, let us first
rebuild the Transformer network from above and then wrap it as a
Creator.

    \begin{tcolorbox}[breakable, size=fbox, boxrule=1pt, pad at break*=1mm,colback=cellbackground, colframe=cellborder]
\prompt{In}{incolor}{ }{\boxspacing}
\begin{Verbatim}[commandchars=\\\{\}]
\PY{k+kn}{from} \PY{n+nn}{deepvoxnet2}\PY{n+nn}{.}\PY{n+nn}{components}\PY{n+nn}{.}\PY{n+nn}{transformers} \PY{k+kn}{import} \PY{n}{MircInput}
\PY{k+kn}{from} \PY{n+nn}{deepvoxnet2}\PY{n+nn}{.}\PY{n+nn}{components}\PY{n+nn}{.}\PY{n+nn}{creator} \PY{k+kn}{import} \PY{n}{Creator}

\PY{n}{x\PYZus{}y} \PY{o}{=} \PY{n}{MircInput}\PY{p}{(}\PY{p}{[}\PY{l+s+s2}{\PYZdq{}}\PY{l+s+s2}{flair}\PY{l+s+s2}{\PYZdq{}}\PY{p}{,} \PY{l+s+s2}{\PYZdq{}}\PY{l+s+s2}{gt}\PY{l+s+s2}{\PYZdq{}}\PY{p}{]}\PY{p}{,} \PY{n}{n}\PY{o}{=}\PY{l+m+mi}{1}\PY{p}{,} \PY{n}{output\PYZus{}shapes}\PY{o}{=}\PY{p}{[}\PY{p}{(}\PY{l+m+mi}{1}\PY{p}{,} \PY{l+m+mi}{240}\PY{p}{,} \PY{l+m+mi}{240}\PY{p}{,} \PY{k+kc}{None}\PY{p}{,} \PY{l+m+mi}{1}\PY{p}{)}\PY{p}{,} \PY{p}{(}\PY{l+m+mi}{1}\PY{p}{,} \PY{l+m+mi}{240}\PY{p}{,} \PY{l+m+mi}{240}\PY{p}{,} \PY{k+kc}{None}\PY{p}{,} \PY{l+m+mi}{1}\PY{p}{)}\PY{p}{]}\PY{p}{)}
\PY{n}{x} \PY{o}{=} \PY{n}{Split}\PY{p}{(}\PY{n}{indices}\PY{o}{=}\PY{p}{(}\PY{l+m+mi}{0}\PY{p}{,}\PY{p}{)}\PY{p}{,} \PY{n}{n}\PY{o}{=}\PY{l+m+mi}{1}\PY{p}{)}\PY{p}{(}\PY{n}{x\PYZus{}y}\PY{p}{)}
\PY{n}{y} \PY{o}{=} \PY{n}{Split}\PY{p}{(}\PY{n}{indices}\PY{o}{=}\PY{p}{(}\PY{l+m+mi}{1}\PY{p}{,}\PY{p}{)}\PY{p}{,} \PY{n}{n}\PY{o}{=}\PY{l+m+mi}{1}\PY{p}{)}\PY{p}{(}\PY{n}{x\PYZus{}y}\PY{p}{)}
\PY{n}{x\PYZus{}creator} \PY{o}{=} \PY{n}{Creator}\PY{p}{(}\PY{n}{outputs}\PY{o}{=}\PY{p}{[}\PY{n}{x}\PY{p}{]}\PY{p}{)}
\PY{n}{y\PYZus{}creator} \PY{o}{=} \PY{n}{Creator}\PY{p}{(}\PY{n}{outputs}\PY{o}{=}\PY{p}{[}\PY{n}{y}\PY{p}{]}\PY{p}{)}
\PY{n}{x\PYZus{}y\PYZus{}creator} \PY{o}{=} \PY{n}{Creator}\PY{p}{(}\PY{n}{outputs}\PY{o}{=}\PY{p}{[}\PY{n}{x}\PY{p}{,} \PY{n}{y}\PY{p}{]}\PY{p}{)}

\PY{k}{def} \PY{n+nf}{visualize\PYZus{}creator\PYZus{}outputs}\PY{p}{(}\PY{n}{creator}\PY{p}{,} \PY{n}{identifier}\PY{p}{)}\PY{p}{:}
    \PY{n}{i} \PY{o}{=} \PY{l+m+mi}{0}
    \PY{k}{for} \PY{n}{creator\PYZus{}outputs\PYZus{}i} \PY{o+ow}{in} \PY{n}{creator}\PY{o}{.}\PY{n}{eval}\PY{p}{(}\PY{n}{identifier}\PY{p}{)}\PY{p}{:}
        \PY{n+nb}{print}\PY{p}{(}\PY{l+s+sa}{f}\PY{l+s+s2}{\PYZdq{}}\PY{l+s+s2}{Evaluation }\PY{l+s+si}{\PYZob{}}\PY{n}{i}\PY{l+s+si}{\PYZcb{}}\PY{l+s+s2}{:}\PY{l+s+s2}{\PYZdq{}}\PY{p}{)}
        \PY{k}{for} \PY{n}{j}\PY{p}{,} \PY{n}{transformer\PYZus{}output\PYZus{}i} \PY{o+ow}{in} \PY{n+nb}{enumerate}\PY{p}{(}\PY{n}{creator\PYZus{}outputs\PYZus{}i}\PY{p}{)}\PY{p}{:}
            \PY{n+nb}{print}\PY{p}{(}\PY{l+s+sa}{f}\PY{l+s+s2}{\PYZdq{}}\PY{l+s+se}{\PYZbs{}t}\PY{l+s+s2}{Output of Transformer }\PY{l+s+si}{\PYZob{}}\PY{n}{j}\PY{l+s+si}{\PYZcb{}}\PY{l+s+s2}{:}\PY{l+s+s2}{\PYZdq{}}\PY{p}{)}
            \PY{n}{visualize\PYZus{}list\PYZus{}of\PYZus{}samples}\PY{p}{(}\PY{n}{transformer\PYZus{}output\PYZus{}i}\PY{p}{)}

        \PY{n}{i} \PY{o}{+}\PY{o}{=} \PY{l+m+mi}{1}

    \PY{n+nb}{print}\PY{p}{(}\PY{l+s+sa}{f}\PY{l+s+s2}{\PYZdq{}}\PY{l+s+s2}{Generation stopped after n=}\PY{l+s+si}{\PYZob{}}\PY{n}{i}\PY{l+s+si}{\PYZcb{}}\PY{l+s+s2}{ evaluations}\PY{l+s+s2}{\PYZdq{}}\PY{p}{)}

\PY{n}{visualize\PYZus{}creator\PYZus{}outputs}\PY{p}{(}\PY{n}{x\PYZus{}creator}\PY{p}{,} \PY{n}{my\PYZus{}val\PYZus{}sampler}\PY{p}{[}\PY{l+m+mi}{1}\PY{p}{]}\PY{p}{)}
\PY{n}{visualize\PYZus{}creator\PYZus{}outputs}\PY{p}{(}\PY{n}{y\PYZus{}creator}\PY{p}{,} \PY{n}{my\PYZus{}val\PYZus{}sampler}\PY{p}{[}\PY{l+m+mi}{1}\PY{p}{]}\PY{p}{)}
\PY{n}{visualize\PYZus{}creator\PYZus{}outputs}\PY{p}{(}\PY{n}{x\PYZus{}y\PYZus{}creator}\PY{p}{,} \PY{n}{my\PYZus{}val\PYZus{}sampler}\PY{p}{[}\PY{l+m+mi}{1}\PY{p}{]}\PY{p}{)}
\end{Verbatim}
\end{tcolorbox}

    Notice how the Creator object has the potential to evaluate the output
of multiple Transformer objects at once (we give it a list of
Transformer outputs upon creation). Evaluating the Creator is somewhat
different from evaluating a Transformer output as we did before. When we
evaluated a Transformer output, we would have first needed to make sure
that all InputTransformer objects had loaded the Identifier
\emph{manually}. Then, using \texttt{.eval()} we updated that
particular output of that particular Transformer. However, when using
\texttt{.eval(identifier)} of the Creator object, the Creator will first
reset all Transformer objects and then load the specified Identifier
into all InputTransformer objects it can find in its Transformer
network. Finally, this will return a proper generator object that you can
run until completion (i.e.~as soon as at least one InputTransformer is
depleted).

Another useful, higher-level functionality of the Creator object is
that it traces back which Transformer objects are needed to
compute the requested Transformer outputs. As a result, it will simplify
the Transformer network internally. Furthermore, we can save the entire
Creator as an object and load it for later use using its static
\texttt{save\_creator} and \texttt{load\_creator} methods. Last but not
least, it will give names to the Transformers, and you will be able to
summarize the network in a printout. Note that 
we also specified some input shapes in the code snippet above. This is optional, but the summary will become more informative when doing so, and you can keep
better track of the shapes.

    \begin{tcolorbox}[breakable, size=fbox, boxrule=1pt, pad at break*=1mm,colback=cellbackground, colframe=cellborder]
\prompt{In}{incolor}{ }{\boxspacing}
\begin{Verbatim}[commandchars=\\\{\}]
\PY{k+kn}{import} \PY{n+nn}{tempfile}

\PY{n}{tmp\PYZus{}dir} \PY{o}{=} \PY{n}{tempfile}\PY{o}{.}\PY{n}{TemporaryDirectory}\PY{p}{(}\PY{p}{)}
\PY{n}{tmp\PYZus{}file} \PY{o}{=} \PY{n}{os}\PY{o}{.}\PY{n}{path}\PY{o}{.}\PY{n}{join}\PY{p}{(}\PY{n}{tmp\PYZus{}dir}\PY{o}{.}\PY{n}{name}\PY{p}{,} \PY{l+s+s2}{\PYZdq{}}\PY{l+s+s2}{x\PYZus{}y\PYZus{}creator.pkl}\PY{l+s+s2}{\PYZdq{}}\PY{p}{)}
\PY{n}{Creator}\PY{o}{.}\PY{n}{save\PYZus{}creator}\PY{p}{(}\PY{n}{x\PYZus{}y\PYZus{}creator}\PY{p}{,} \PY{n}{file\PYZus{}path}\PY{o}{=}\PY{n}{tmp\PYZus{}file}\PY{p}{)}  \PY{c+c1}{\PYZsh{} or x\PYZus{}y\PYZus{}creator.save(tmp\PYZus{}file)}
\PY{n}{x\PYZus{}y\PYZus{}creator} \PY{o}{=} \PY{n}{Creator}\PY{o}{.}\PY{n}{load\PYZus{}creator}\PY{p}{(}\PY{n}{tmp\PYZus{}file}\PY{p}{)}
\PY{n}{x\PYZus{}y\PYZus{}creator}\PY{o}{.}\PY{n}{summary}\PY{p}{(}\PY{p}{)}
\PY{n}{tmp\PYZus{}dir}\PY{o}{.}\PY{n}{cleanup}\PY{p}{(}\PY{p}{)}
\end{Verbatim}
\end{tcolorbox}

    Be careful with saving and loading Creator objects as soon as you have
KerasModel Transformer objects in your network of Transformer objects
(see later in Section~\ref{the-kerasmodel-transformer}). The Keras models will be cleared before
saving; thus, you should manually save, load and set them using the
\texttt{set\_keras\_models} method. As soon as you use a DvnModel (Section~\ref{the-dvnmodel-object}) and use its own save and load methods, this will be taken care
of for you.

    \subsection{More complex Transformer
networks}\label{more-complex-transformer-networks}

The beauty of the Transformer network is that the transformation can
become arbitrarily complex, while the construction of it remains
relatively simple. Without describing the functioning of all available
Transformer objects, there are a couple of intricacies that are still
useful to explain.

The first thing has to do with the data pair, i.e.~a list of Sample
objects, and was already part of the above example code. 
To be clear, it is important to not view a data pair as a pair but
rather as a list containing one or more Sample objects. This pair does not need to remain
intact when flowing through the Transformer network. As a result, the resulting set may look entirely different when applying a sampling function (i.e.~combination
of the Sampler and Transformer network). After the use of the Split Transformer, we could use the
Group Transformer to group again, resulting in a set 
identical to the original (depending on the Sampler, of course).

    \begin{tcolorbox}[breakable, size=fbox, boxrule=1pt, pad at break*=1mm,colback=cellbackground, colframe=cellborder]
\prompt{In}{incolor}{ }{\boxspacing}
\begin{Verbatim}[commandchars=\\\{\}]
\PY{k+kn}{from} \PY{n+nn}{deepvoxnet2}\PY{n+nn}{.}\PY{n+nn}{components}\PY{n+nn}{.}\PY{n+nn}{transformers} \PY{k+kn}{import} \PY{n}{Group}

\PY{n}{x\PYZus{}y} \PY{o}{=} \PY{n}{Group}\PY{p}{(}\PY{p}{)}\PY{p}{(}\PY{p}{[}\PY{n}{x}\PY{p}{,} \PY{n}{y}\PY{p}{]}\PY{p}{)}
\PY{n}{x\PYZus{}y\PYZus{}creator} \PY{o}{=} \PY{n}{Creator}\PY{p}{(}\PY{n}{outputs}\PY{o}{=}\PY{p}{[}\PY{n}{x\PYZus{}y}\PY{p}{]}\PY{p}{)}
\PY{n}{visualize\PYZus{}creator\PYZus{}outputs}\PY{p}{(}\PY{n}{x\PYZus{}y\PYZus{}creator}\PY{p}{,} \PY{n}{my\PYZus{}val\PYZus{}sampler}\PY{p}{[}\PY{l+m+mi}{1}\PY{p}{]}\PY{p}{)}
\end{Verbatim}
\end{tcolorbox}

    Also, note that you can use multiple InputTransformers and that they will
see the same Identifier. This can be useful since it allows you to build
one network and wrap different parts of it in different Creator objects.
Since the Creator simplifies the underlying Transformer network, it is
possible that not all InputTransformers are used and thus, you could
evaluate certain parts with an Identifier lacking certain information.
Just as an example how we could use two InputTransformer objects
instead:

    \begin{tcolorbox}[breakable, size=fbox, boxrule=1pt, pad at break*=1mm,colback=cellbackground, colframe=cellborder]
\prompt{In}{incolor}{ }{\boxspacing}
\begin{Verbatim}[commandchars=\\\{\}]
\PY{n}{x} \PY{o}{=} \PY{n}{x\PYZus{}input} \PY{o}{=} \PY{n}{MircInput}\PY{p}{(}\PY{p}{[}\PY{l+s+s2}{\PYZdq{}}\PY{l+s+s2}{flair}\PY{l+s+s2}{\PYZdq{}}\PY{p}{]}\PY{p}{,} \PY{n}{n}\PY{o}{=}\PY{l+m+mi}{1}\PY{p}{,} \PY{n}{output\PYZus{}shapes}\PY{o}{=}\PY{p}{[}\PY{p}{(}\PY{l+m+mi}{1}\PY{p}{,} \PY{l+m+mi}{240}\PY{p}{,} \PY{l+m+mi}{240}\PY{p}{,} \PY{k+kc}{None}\PY{p}{,} \PY{l+m+mi}{1}\PY{p}{)}\PY{p}{]}\PY{p}{)}
\PY{n}{y} \PY{o}{=} \PY{n}{y\PYZus{}input} \PY{o}{=} \PY{n}{MircInput}\PY{p}{(}\PY{p}{[}\PY{l+s+s2}{\PYZdq{}}\PY{l+s+s2}{gt}\PY{l+s+s2}{\PYZdq{}}\PY{p}{]}\PY{p}{,} \PY{n}{n}\PY{o}{=}\PY{l+m+mi}{1}\PY{p}{,} \PY{n}{output\PYZus{}shapes}\PY{o}{=}\PY{p}{[}\PY{p}{(}\PY{l+m+mi}{1}\PY{p}{,} \PY{l+m+mi}{240}\PY{p}{,} \PY{l+m+mi}{240}\PY{p}{,} \PY{k+kc}{None}\PY{p}{,} \PY{l+m+mi}{1}\PY{p}{)}\PY{p}{]}\PY{p}{)}
\PY{n}{x\PYZus{}y} \PY{o}{=} \PY{n}{Group}\PY{p}{(}\PY{p}{)}\PY{p}{(}\PY{p}{[}\PY{n}{x}\PY{p}{,} \PY{n}{y}\PY{p}{]}\PY{p}{)}
\PY{n}{x\PYZus{}y\PYZus{}creator} \PY{o}{=} \PY{n}{Creator}\PY{p}{(}\PY{n}{outputs}\PY{o}{=}\PY{p}{[}\PY{n}{x\PYZus{}y}\PY{p}{]}\PY{p}{)}
\PY{n}{visualize\PYZus{}creator\PYZus{}outputs}\PY{p}{(}\PY{n}{x\PYZus{}y\PYZus{}creator}\PY{p}{,} \PY{n}{my\PYZus{}val\PYZus{}sampler}\PY{p}{[}\PY{l+m+mi}{1}\PY{p}{]}\PY{p}{)}
\end{Verbatim}
\end{tcolorbox}

    Second, some Transformer objects need a reference Transformer to
calculate their internal transformation. For example, if we want to
deform the Samples using an affine transformation, we want first to compute the affine transform based on a reference Sample. Similarly, if
we want to produce some crops, we might want those to originate from
non-zero portions of a particular Sample. For this purpose, it is always a
good idea to have the reference output be a list of only one Sample,
such that there is no ambiguity in which Sample is used as a reference.

    \begin{tcolorbox}[breakable, size=fbox, boxrule=1pt, pad at break*=1mm,colback=cellbackground, colframe=cellborder]
\prompt{In}{incolor}{ }{\boxspacing}
\begin{Verbatim}[commandchars=\\\{\}]
\PY{k+kn}{from} \PY{n+nn}{deepvoxnet2}\PY{n+nn}{.}\PY{n+nn}{components}\PY{n+nn}{.}\PY{n+nn}{transformers} \PY{k+kn}{import} \PY{n}{AffineDeformation}\PY{p}{,} \PY{n}{RandomCrop}\PY{p}{,} \PY{n}{Threshold}\PY{p}{,} \PY{n}{Flip}

\PY{n}{x\PYZus{}y\PYZus{}affine} \PY{o}{=} \PY{n}{AffineDeformation}\PY{p}{(}\PY{n}{x}\PY{p}{,} \PY{n}{rotation\PYZus{}window\PYZus{}width}\PY{o}{=}\PY{p}{(}\PY{l+m+mi}{1}\PY{p}{,} \PY{l+m+mi}{0}\PY{p}{,} \PY{l+m+mi}{0}\PY{p}{)}\PY{p}{,} \PY{n}{translation\PYZus{}window\PYZus{}width}\PY{o}{=}\PY{p}{(}\PY{l+m+mi}{10}\PY{p}{,} \PY{l+m+mi}{10}\PY{p}{,} \PY{l+m+mi}{0}\PY{p}{)}\PY{p}{)}\PY{p}{(}\PY{n}{x\PYZus{}y}\PY{p}{)}
\PY{n}{x\PYZus{}y\PYZus{}flip} \PY{o}{=} \PY{n}{Flip}\PY{p}{(}\PY{n}{flip\PYZus{}probabilities}\PY{o}{=}\PY{p}{(}\PY{l+m+mf}{0.5}\PY{p}{,} \PY{l+m+mi}{0}\PY{p}{,} \PY{l+m+mi}{0}\PY{p}{)}\PY{p}{,} \PY{n}{n}\PY{o}{=}\PY{l+m+mi}{2}\PY{p}{)}\PY{p}{(}\PY{n}{x\PYZus{}y\PYZus{}affine}\PY{p}{)}
\PY{n}{x\PYZus{}flip} \PY{o}{=} \PY{n}{Split}\PY{p}{(}\PY{n}{indices}\PY{o}{=}\PY{p}{(}\PY{l+m+mi}{0}\PY{p}{,}\PY{p}{)}\PY{p}{)}\PY{p}{(}\PY{n}{x\PYZus{}y\PYZus{}flip}\PY{p}{)}
\PY{n}{mask\PYZus{}flip} \PY{o}{=} \PY{n}{Threshold}\PY{p}{(}\PY{n}{lower\PYZus{}threshold}\PY{o}{=}\PY{l+m+mi}{0}\PY{p}{)}\PY{p}{(}\PY{n}{x\PYZus{}flip}\PY{p}{)}
\PY{n}{x\PYZus{}y\PYZus{}crop} \PY{o}{=} \PY{n}{RandomCrop}\PY{p}{(}\PY{n}{mask\PYZus{}flip}\PY{p}{,} \PY{p}{(}\PY{l+m+mi}{85}\PY{p}{,} \PY{l+m+mi}{85}\PY{p}{,} \PY{l+m+mi}{85}\PY{p}{)}\PY{p}{,} \PY{n}{nonzero}\PY{o}{=}\PY{k+kc}{True}\PY{p}{,} \PY{n}{n}\PY{o}{=}\PY{l+m+mi}{4}\PY{p}{)}\PY{p}{(}\PY{n}{x\PYZus{}y\PYZus{}flip}\PY{p}{)}
\PY{n}{x\PYZus{}y\PYZus{}crop\PYZus{}creator} \PY{o}{=} \PY{n}{Creator}\PY{p}{(}\PY{n}{outputs}\PY{o}{=}\PY{p}{[}\PY{n}{x\PYZus{}y\PYZus{}crop}\PY{p}{]}\PY{p}{)}
\PY{n}{visualize\PYZus{}creator\PYZus{}outputs}\PY{p}{(}\PY{n}{x\PYZus{}y\PYZus{}crop\PYZus{}creator}\PY{p}{,} \PY{n}{my\PYZus{}val\PYZus{}sampler}\PY{p}{[}\PY{l+m+mi}{1}\PY{p}{]}\PY{p}{)}
\PY{n}{x\PYZus{}y\PYZus{}crop\PYZus{}creator\PYZus{}} \PY{o}{=} \PY{n}{Creator}\PY{p}{(}\PY{n}{outputs}\PY{o}{=}\PY{p}{[}\PY{n}{x\PYZus{}y\PYZus{}affine}\PY{p}{,} \PY{n}{x\PYZus{}y\PYZus{}flip}\PY{p}{,} \PY{n}{mask\PYZus{}flip}\PY{p}{,} \PY{n}{x\PYZus{}y\PYZus{}crop}\PY{p}{]}\PY{p}{)}
\PY{n}{visualize\PYZus{}creator\PYZus{}outputs}\PY{p}{(}\PY{n}{x\PYZus{}y\PYZus{}crop\PYZus{}creator\PYZus{}}\PY{p}{,} \PY{n}{my\PYZus{}val\PYZus{}sampler}\PY{p}{[}\PY{l+m+mi}{1}\PY{p}{]}\PY{p}{)}
\end{Verbatim}
\end{tcolorbox}

    Notice how \emph{x\_y\_crop\_creator} generates N=1*1*1*1*2*4=8 crops
while the \emph{x\_y\_crop\_creator\_} only generates a single crop.
This is because the \emph{x\_y\_crop\_creator\_} also generates
the output of the AffineDeformation Transformer, and requesting it to
generate an output the second time will deplete the underlying
InputTransformer objects already (\texttt{n=1} for most Transformer
objects by default). So be careful with the so-called \emph{shortcut
connections}.

    Third, remember that we said that the input and output of a Transformer
is a list of Sample objects. In fact, this is only partly true and this
has to do with the concept that a Transformer is applying the same
random transformation to all Sample objects it can find at its input. A
Transformer can be connected to multiple inputs, resulting in multiple
outputs, meaning that a Transformer may transform multiple data pairs
that are connected to the Transformer. As such, we could have
constructed the Transformer network from above in many different ways,
so always choose the most efficient route. Try to find some differences
in the below code:

    \begin{tcolorbox}[breakable, size=fbox, boxrule=1pt, pad at break*=1mm,colback=cellbackground, colframe=cellborder]
\prompt{In}{incolor}{ }{\boxspacing}
\begin{Verbatim}[commandchars=\\\{\}]
\PY{n}{affine\PYZus{}transformer} \PY{o}{=} \PY{n}{AffineDeformation}\PY{p}{(}\PY{n}{x}\PY{p}{,} \PY{n}{rotation\PYZus{}window\PYZus{}width}\PY{o}{=}\PY{p}{(}\PY{l+m+mi}{1}\PY{p}{,} \PY{l+m+mi}{0}\PY{p}{,} \PY{l+m+mi}{0}\PY{p}{)}\PY{p}{,} \PY{n}{translation\PYZus{}window\PYZus{}width}\PY{o}{=}\PY{p}{(}\PY{l+m+mi}{10}\PY{p}{,} \PY{l+m+mi}{10}\PY{p}{,} \PY{l+m+mi}{0}\PY{p}{)}\PY{p}{)}
\PY{n}{x\PYZus{}affine}\PY{p}{,} \PY{n}{y\PYZus{}affine} \PY{o}{=} \PY{n}{affine\PYZus{}transformer}\PY{p}{(}\PY{n}{x\PYZus{}input}\PY{p}{,} \PY{n}{y\PYZus{}input}\PY{p}{)}
\PY{n}{flip\PYZus{}transformer} \PY{o}{=} \PY{n}{Flip}\PY{p}{(}\PY{n}{flip\PYZus{}probabilities}\PY{o}{=}\PY{p}{(}\PY{l+m+mf}{0.5}\PY{p}{,} \PY{l+m+mi}{0}\PY{p}{,} \PY{l+m+mi}{0}\PY{p}{)}\PY{p}{,} \PY{n}{n}\PY{o}{=}\PY{l+m+mi}{2}\PY{p}{)}
\PY{n}{x\PYZus{}flip} \PY{o}{=} \PY{n}{flip\PYZus{}transformer}\PY{p}{(}\PY{n}{x\PYZus{}affine}\PY{p}{)}
\PY{n}{y\PYZus{}flip} \PY{o}{=} \PY{n}{flip\PYZus{}transformer}\PY{p}{(}\PY{n}{y\PYZus{}affine}\PY{p}{)}
\PY{n}{mask\PYZus{}flip} \PY{o}{=} \PY{n}{Threshold}\PY{p}{(}\PY{n}{lower\PYZus{}threshold}\PY{o}{=}\PY{l+m+mi}{0}\PY{p}{)}\PY{p}{(}\PY{n}{x\PYZus{}flip}\PY{p}{)}
\PY{n}{x\PYZus{}crop}\PY{p}{,} \PY{n}{y\PYZus{}crop} \PY{o}{=} \PY{n}{RandomCrop}\PY{p}{(}\PY{n}{mask\PYZus{}flip}\PY{p}{,} \PY{p}{(}\PY{l+m+mi}{85}\PY{p}{,} \PY{l+m+mi}{85}\PY{p}{,} \PY{l+m+mi}{85}\PY{p}{)}\PY{p}{,} \PY{n}{nonzero}\PY{o}{=}\PY{k+kc}{True}\PY{p}{,} \PY{n}{n}\PY{o}{=}\PY{l+m+mi}{4}\PY{p}{)}\PY{p}{(}\PY{n}{x\PYZus{}flip}\PY{p}{,} \PY{n}{y\PYZus{}flip}\PY{p}{)}
\PY{n}{x\PYZus{}y\PYZus{}crop} \PY{o}{=} \PY{n}{Group}\PY{p}{(}\PY{p}{)}\PY{p}{(}\PY{p}{[}\PY{n}{x\PYZus{}crop}\PY{p}{,} \PY{n}{y\PYZus{}crop}\PY{p}{]}\PY{p}{)}
\PY{n}{x\PYZus{}y\PYZus{}crop\PYZus{}creator} \PY{o}{=} \PY{n}{Creator}\PY{p}{(}\PY{n}{outputs}\PY{o}{=}\PY{p}{[}\PY{n}{x\PYZus{}y\PYZus{}crop}\PY{p}{]}\PY{p}{)}
\PY{n}{visualize\PYZus{}creator\PYZus{}outputs}\PY{p}{(}\PY{n}{x\PYZus{}y\PYZus{}crop\PYZus{}creator}\PY{p}{,} \PY{n}{my\PYZus{}val\PYZus{}sampler}\PY{p}{[}\PY{l+m+mi}{1}\PY{p}{]}\PY{p}{)}
\end{Verbatim}
\end{tcolorbox}

    Again, notice how the above Transformer network can generate 8
transformed data pairs starting from a particular identifier. As soon as
the network gets more complex, it is always a good idea if the
generated number corresponds to what you had in mind. It is possible
that an InputTransformer triggers to stop the generation too soon due to
unwanted shortcuts in your network, e.g.~your network contains parallel
paths with one path generating more than the other.

Finally, the most remarkable thing about the Transformer network is that it
works with Sample objects, hence it always has a notion of where the
Sample is located in the \emph{world}. We could use the Put Transformer
to \emph{put back} all the image crops into a reference Sample space of
choice.

    \begin{tcolorbox}[breakable, size=fbox, boxrule=1pt, pad at break*=1mm,colback=cellbackground, colframe=cellborder]
\prompt{In}{incolor}{ }{\boxspacing}
\begin{Verbatim}[commandchars=\\\{\}]
\PY{k+kn}{from} \PY{n+nn}{deepvoxnet2}\PY{n+nn}{.}\PY{n+nn}{components}\PY{n+nn}{.}\PY{n+nn}{transformers} \PY{k+kn}{import} \PY{n}{Buffer}\PY{p}{,} \PY{n}{Put}

\PY{n}{x\PYZus{}buffer} \PY{o}{=} \PY{n}{Buffer}\PY{p}{(}\PY{n}{buffer\PYZus{}size}\PY{o}{=}\PY{k+kc}{None}\PY{p}{)}\PY{p}{(}\PY{n}{x\PYZus{}crop}\PY{p}{)}
\PY{n}{x\PYZus{}put} \PY{o}{=} \PY{n}{Put}\PY{p}{(}\PY{n}{reference\PYZus{}connection}\PY{o}{=}\PY{n}{x\PYZus{}input}\PY{p}{)}\PY{p}{(}\PY{n}{x\PYZus{}buffer}\PY{p}{)}
\PY{n}{x\PYZus{}put\PYZus{}creator} \PY{o}{=} \PY{n}{Creator}\PY{p}{(}\PY{n}{outputs}\PY{o}{=}\PY{p}{[}\PY{n}{x\PYZus{}put}\PY{p}{]}\PY{p}{)}
\PY{n}{visualize\PYZus{}creator\PYZus{}outputs}\PY{p}{(}\PY{n}{x\PYZus{}put\PYZus{}creator}\PY{p}{,} \PY{n}{my\PYZus{}val\PYZus{}sampler}\PY{p}{[}\PY{l+m+mi}{1}\PY{p}{]}\PY{p}{)}
\end{Verbatim}
\end{tcolorbox}

    As an exercise, you can explore what would happen with the reconstructed
image if you would take more crops (increase n of RandomCrop).
Alternatively, what happens if you use the GridCrop instead of the
RandomCrop using the default \texttt{n=None} option?

    \section{DVN2 for CNN-based
applications}\label{dvn2-for-cnn-based-applications}

At this point, we can generate, albeit still implicitly, a wide variety
of sets, e.g.~\(\mathcal{S''}\), starting from the set \(S'\). Using the
suggested \gls{dvn2} way, we could operate as follows: 

\begin{itemize}
    \tightlist
    \item organize the set of data pairs \(\mathcal{S'}\) as Record objects in a Mirc object;
    \item create a MircSampler from the Mirc object to sample MircIdentifiers;
    \item make a Transformer network (as a Creator) that starts from a MircIdentifier and generates new data pairs.
\end{itemize}

As a result, we have a powerful tool at our disposal that we can use to
make \gls{cnn}-based application pipelines. In order to effectively use it, in
\gls{dvn2} we linked with the Tensorflow library, particularly Keras.
Nevertheless, until this point, everything was coded in Python independently of the deep learning framework. The
following functions can be adapted when desired to use other deep learning
frameworks such as Pytorch.

    \subsection{The KerasModel
Transformer}\label{the-kerasmodel-transformer}

First of all, in \gls{dvn2} we have provided functions to create a DeepMedic-
or U-Net-like \gls{cnn} architecture as a Keras model. It provides a variety
of options, such as different types of layer normalizations, residual
connections, siam networks, padding, etc. Furthermore, it provides the
user with plenty of information, for example, the \gls{rf} and
possible output sizes, and even an estimated memory usage. When using
the default settings, it will create the No-New Net variant of the
standard U-Net:

    \begin{tcolorbox}[breakable, size=fbox, boxrule=1pt, pad at break*=1mm,colback=cellbackground, colframe=cellborder]
\prompt{In}{incolor}{ }{\boxspacing}
\begin{Verbatim}[commandchars=\\\{\}]
\PY{k+kn}{from} \PY{n+nn}{deepvoxnet2}\PY{n+nn}{.}\PY{n+nn}{keras}\PY{n+nn}{.}\PY{n+nn}{models}\PY{n+nn}{.}\PY{n+nn}{unet\PYZus{}generalized\PYZus{}v2} \PY{k+kn}{import} \PY{n}{create\PYZus{}generalized\PYZus{}unet\PYZus{}v2\PYZus{}model}

\PY{n}{no\PYZus{}new\PYZus{}net\PYZus{}model} \PY{o}{=} \PY{n}{create\PYZus{}generalized\PYZus{}unet\PYZus{}v2\PYZus{}model}\PY{p}{(}\PY{p}{)}
\end{Verbatim}
\end{tcolorbox}

    We can see that the model performs internal padding because both the
output and input size is 128 x~128 x~128~voxels\textsuperscript{3}. We also see that the \gls{rf} size is 185 x~185 x~185~voxels\textsuperscript{3}. Besides plenty of other information, the
summary provides us with potential alternative output and input sizes.
Suppose we want to create a simple U-Net-like \gls{cnn} that transforms a
single input patch of size 85 x~85 x~85~voxels\textsuperscript{3} into an output patch of 53 x~53 x~53~voxels\textsuperscript{3},
representing the predicted segmentation. To do so, we could modify some
of the default options like below:

    \begin{tcolorbox}[breakable, size=fbox, boxrule=1pt, pad at break*=1mm,colback=cellbackground, colframe=cellborder]
\prompt{In}{incolor}{ }{\boxspacing}
\begin{Verbatim}[commandchars=\\\{\}]
\PY{n}{my\PYZus{}own\PYZus{}unet\PYZus{}model} \PY{o}{=} \PY{n}{create\PYZus{}generalized\PYZus{}unet\PYZus{}v2\PYZus{}model}\PY{p}{(}
    \PY{n}{number\PYZus{}input\PYZus{}features}\PY{o}{=}\PY{l+m+mi}{1}\PY{p}{,}
    \PY{n}{subsample\PYZus{}factors\PYZus{}per\PYZus{}pathway}\PY{o}{=}\PY{p}{(}
            \PY{p}{(}\PY{l+m+mi}{1}\PY{p}{,} \PY{l+m+mi}{1}\PY{p}{,} \PY{l+m+mi}{1}\PY{p}{)}\PY{p}{,}
            \PY{p}{(}\PY{l+m+mi}{3}\PY{p}{,} \PY{l+m+mi}{3}\PY{p}{,} \PY{l+m+mi}{3}\PY{p}{)}
    \PY{p}{)}\PY{p}{,}
    \PY{n}{kernel\PYZus{}sizes\PYZus{}per\PYZus{}pathway}\PY{o}{=}\PY{p}{(}
            \PY{p}{(}\PY{p}{(}\PY{p}{(}\PY{l+m+mi}{3}\PY{p}{,} \PY{l+m+mi}{3}\PY{p}{,} \PY{l+m+mi}{3}\PY{p}{)}\PY{p}{,} \PY{p}{(}\PY{l+m+mi}{3}\PY{p}{,} \PY{l+m+mi}{3}\PY{p}{,} \PY{l+m+mi}{3}\PY{p}{)}\PY{p}{)}\PY{p}{,} \PY{p}{(}\PY{p}{(}\PY{l+m+mi}{3}\PY{p}{,} \PY{l+m+mi}{3}\PY{p}{,} \PY{l+m+mi}{3}\PY{p}{)}\PY{p}{,} \PY{p}{(}\PY{l+m+mi}{3}\PY{p}{,} \PY{l+m+mi}{3}\PY{p}{,} \PY{l+m+mi}{3}\PY{p}{)}\PY{p}{)}\PY{p}{)}\PY{p}{,}
            \PY{p}{(}\PY{p}{(}\PY{p}{(}\PY{l+m+mi}{3}\PY{p}{,} \PY{l+m+mi}{3}\PY{p}{,} \PY{l+m+mi}{3}\PY{p}{)}\PY{p}{,} \PY{p}{(}\PY{l+m+mi}{3}\PY{p}{,} \PY{l+m+mi}{3}\PY{p}{,} \PY{l+m+mi}{3}\PY{p}{)}\PY{p}{)}\PY{p}{,} \PY{p}{(}\PY{p}{(}\PY{l+m+mi}{3}\PY{p}{,} \PY{l+m+mi}{3}\PY{p}{,} \PY{l+m+mi}{3}\PY{p}{)}\PY{p}{,} \PY{p}{(}\PY{l+m+mi}{3}\PY{p}{,} \PY{l+m+mi}{3}\PY{p}{,} \PY{l+m+mi}{3}\PY{p}{)}\PY{p}{)}\PY{p}{)}
    \PY{p}{)}\PY{p}{,}
    \PY{n}{number\PYZus{}features\PYZus{}per\PYZus{}pathway}\PY{o}{=}\PY{p}{(}
            \PY{p}{(}\PY{p}{(}\PY{l+m+mi}{30}\PY{p}{,} \PY{l+m+mi}{30}\PY{p}{)}\PY{p}{,} \PY{p}{(}\PY{l+m+mi}{30}\PY{p}{,} \PY{l+m+mi}{30}\PY{p}{)}\PY{p}{)}\PY{p}{,}
            \PY{p}{(}\PY{p}{(}\PY{l+m+mi}{60}\PY{p}{,} \PY{l+m+mi}{60}\PY{p}{)}\PY{p}{,} \PY{p}{(}\PY{l+m+mi}{60}\PY{p}{,} \PY{l+m+mi}{30}\PY{p}{)}\PY{p}{)}
    \PY{p}{)}\PY{p}{,}
    \PY{n}{output\PYZus{}size}\PY{o}{=}\PY{p}{(}\PY{l+m+mi}{53}\PY{p}{,} \PY{l+m+mi}{53}\PY{p}{,} \PY{l+m+mi}{53}\PY{p}{)}\PY{p}{,}
    \PY{n}{padding}\PY{o}{=}\PY{l+s+s2}{\PYZdq{}}\PY{l+s+s2}{valid}\PY{l+s+s2}{\PYZdq{}}\PY{p}{,}
    \PY{n}{instance\PYZus{}normalization}\PY{o}{=}\PY{k+kc}{False}\PY{p}{,}
    \PY{n}{batch\PYZus{}normalization}\PY{o}{=}\PY{k+kc}{True}
\PY{p}{)}
\end{Verbatim}
\end{tcolorbox}

    In order to proceed, it is important to understand that a \gls{cnn} can be
seen as a Transformer. The standard view of a \gls{cnn} is that it transforms
inputs into outputs, specifically, a list of arrays into another
list of arrays. Or, sometimes viewed as transforming a specific (joint)
distribution into another (joint) distribution. Unique to \gls{dvn2} is that we
keep this view on \gls{cnn}s but add the spatiality to it, thus transforming
a specific list of Sample objects into another. To do so, \gls{dvn2} provides
the KerasModel Transformer, which wraps a \gls{cnn} as a Transformer
that you can use in your Transformer network. Using the default options,
it assumes the output Sample objects have the same scale as the first
input Sample object and are spatially centered. If this is not
the case, you can use the \texttt{output\_affines} and
\texttt{output\_to\_input} options to specify the affine transformations
from output to input and the input reference for each output,
respectively.

    \begin{tcolorbox}[breakable, size=fbox, boxrule=1pt, pad at break*=1mm,colback=cellbackground, colframe=cellborder]
\prompt{In}{incolor}{ }{\boxspacing}
\begin{Verbatim}[commandchars=\\\{\}]
\PY{k+kn}{from} \PY{n+nn}{deepvoxnet2}\PY{n+nn}{.}\PY{n+nn}{components}\PY{n+nn}{.}\PY{n+nn}{transformers} \PY{k+kn}{import} \PY{n}{KerasModel}

\PY{n}{x\PYZus{}crop\PYZus{}pred} \PY{o}{=} \PY{n}{KerasModel}\PY{p}{(}\PY{n}{my\PYZus{}own\PYZus{}unet\PYZus{}model}\PY{p}{)}\PY{p}{(}\PY{n}{x\PYZus{}crop}\PY{p}{)}
\PY{n}{x\PYZus{}buffer} \PY{o}{=} \PY{n}{Buffer}\PY{p}{(}\PY{n}{buffer\PYZus{}size}\PY{o}{=}\PY{k+kc}{None}\PY{p}{)}\PY{p}{(}\PY{n}{x\PYZus{}crop\PYZus{}pred}\PY{p}{)}
\PY{n}{x\PYZus{}put} \PY{o}{=} \PY{n}{Put}\PY{p}{(}\PY{n}{reference\PYZus{}connection}\PY{o}{=}\PY{n}{x\PYZus{}input}\PY{p}{)}\PY{p}{(}\PY{n}{x\PYZus{}buffer}\PY{p}{)}
\PY{n}{x\PYZus{}put\PYZus{}creator} \PY{o}{=} \PY{n}{Creator}\PY{p}{(}\PY{n}{outputs}\PY{o}{=}\PY{p}{[}\PY{n}{x\PYZus{}put}\PY{p}{]}\PY{p}{)}
\PY{n}{visualize\PYZus{}creator\PYZus{}outputs}\PY{p}{(}\PY{n}{x\PYZus{}put\PYZus{}creator}\PY{p}{,} \PY{n}{my\PYZus{}val\PYZus{}sampler}\PY{p}{[}\PY{l+m+mi}{1}\PY{p}{]}\PY{p}{)}
\end{Verbatim}
\end{tcolorbox}

    Note how we have reconstructed a prediction by letting the crop flow
through the \gls{cnn} and putting it back in the original image space. At this
point, the \gls{cnn} is not trained and the weights are still random. How can
we tune these weights?

\subsection{Tensorflow Dataset
wrapper}\label{tensorflow-dataset-wrapper}

We only need to define the set of data
pairs that contain learning pairs that can be used to train the \gls{cnn}. We
can identify the proper Transformer objects, put them in a Creator
and make a true set of data pairs together with a Sampler. The Tensorflow Dataset is used for this purpose, which is just another
representation for a set of data pairs. The combination of a Creator and
a Sampler has all information that is necessary to construct a
Tensorflow Dataset.

    \begin{tcolorbox}[breakable, size=fbox, boxrule=1pt, pad at break*=1mm,colback=cellbackground, colframe=cellborder]
\prompt{In}{incolor}{ }{\boxspacing}
\begin{Verbatim}[commandchars=\\\{\}]
\PY{k+kn}{import} \PY{n+nn}{time}
\PY{k+kn}{from} \PY{n+nn}{deepvoxnet2}\PY{n+nn}{.}\PY{n+nn}{components}\PY{n+nn}{.}\PY{n+nn}{transformers} \PY{k+kn}{import} \PY{n}{Crop}
\PY{k+kn}{from} \PY{n+nn}{deepvoxnet2}\PY{n+nn}{.}\PY{n+nn}{components}\PY{n+nn}{.}\PY{n+nn}{model} \PY{k+kn}{import} \PY{n}{TfDataset}

\PY{n}{y\PYZus{}crop\PYZus{}true} \PY{o}{=} \PY{n}{Crop}\PY{p}{(}\PY{n}{y\PYZus{}crop}\PY{p}{,} \PY{p}{(}\PY{l+m+mi}{53}\PY{p}{,} \PY{l+m+mi}{53}\PY{p}{,} \PY{l+m+mi}{53}\PY{p}{)}\PY{p}{)}\PY{p}{(}\PY{n}{y\PYZus{}crop}\PY{p}{)}
\PY{n}{learning\PYZus{}set\PYZus{}creator} \PY{o}{=} \PY{n}{Creator}\PY{p}{(}\PY{p}{[}\PY{n}{x\PYZus{}crop}\PY{p}{,} \PY{n}{y\PYZus{}crop\PYZus{}true}\PY{p}{]}\PY{p}{)}
\PY{n}{my\PYZus{}tf\PYZus{}train\PYZus{}dataset} \PY{o}{=} \PY{n}{TfDataset}\PY{p}{(}\PY{n}{learning\PYZus{}set\PYZus{}creator}\PY{p}{,} \PY{n}{my\PYZus{}train\PYZus{}sampler}\PY{p}{,} \PY{n}{batch\PYZus{}size}\PY{o}{=}\PY{l+m+mi}{1}\PY{p}{)}
\PY{n}{count} \PY{o}{=} \PY{l+m+mi}{0}
\PY{n}{prev\PYZus{}time} \PY{o}{=} \PY{n}{time}\PY{o}{.}\PY{n}{time}\PY{p}{(}\PY{p}{)}
\PY{k}{for} \PY{n}{element} \PY{o+ow}{in} \PY{n}{my\PYZus{}tf\PYZus{}train\PYZus{}dataset}\PY{p}{:}
    \PY{n+nb}{print}\PY{p}{(}\PY{l+s+s2}{\PYZdq{}}\PY{l+s+s2}{Generation took }\PY{l+s+si}{\PYZob{}:.2f\PYZcb{}}\PY{l+s+s2}{ s}\PY{l+s+s2}{\PYZdq{}}\PY{o}{.}\PY{n}{format}\PY{p}{(}\PY{n}{time}\PY{o}{.}\PY{n}{time}\PY{p}{(}\PY{p}{)} \PY{o}{\PYZhy{}} \PY{n}{prev\PYZus{}time}\PY{p}{)}\PY{p}{,} \PY{p}{[}\PY{p}{[}\PY{n}{sample}\PY{o}{.}\PY{n}{shape} \PY{k}{for} \PY{n}{sample} \PY{o+ow}{in} \PY{n}{transformer\PYZus{}output}\PY{p}{]} \PY{k}{for} \PY{n}{transformer\PYZus{}output} \PY{o+ow}{in} \PY{n}{element}\PY{p}{]}\PY{p}{)}
    \PY{n}{count} \PY{o}{+}\PY{o}{=} \PY{l+m+mi}{1}
    \PY{n}{prev\PYZus{}time} \PY{o}{=} \PY{n}{time}\PY{o}{.}\PY{n}{time}\PY{p}{(}\PY{p}{)}

\PY{n+nb}{print}\PY{p}{(}\PY{l+s+sa}{f}\PY{l+s+s2}{\PYZdq{}}\PY{l+s+s2}{The entire my\PYZus{}tf\PYZus{}train\PYZus{}dataset contains }\PY{l+s+si}{\PYZob{}}\PY{n}{count}\PY{l+s+si}{\PYZcb{}}\PY{l+s+s2}{ elements}\PY{l+s+s2}{\PYZdq{}}\PY{p}{)}
\end{Verbatim}
\end{tcolorbox}

    Using the standard options, it iterates through
\emph{my\_train\_sampler} only once. Before going to the next Identifier
in the Sampler, it will ask the underlying Transformer network in the
Creator to run until completion. This is the reason why a total of N=8*8
elements were produced, and that it seemed to go in blocks of 8. Looking
at the structure and shape of an element, this is precisely what we would
expect from our \emph{learning\_set\_creator}. We could use this dataset
to train the \gls{cnn}. With a batch size of 1, each epoch would contain 64 updates, which we could define here as one loop through the entire new set
of transformed data pairs. In order to better shuffle across subjects
(Identifiers, actually) and make the generation quicker, we could play
around with some options.

    \begin{tcolorbox}[breakable, size=fbox, boxrule=1pt, pad at break*=1mm,colback=cellbackground, colframe=cellborder]
\prompt{In}{incolor}{ }{\boxspacing}
\begin{Verbatim}[commandchars=\\\{\}]
\PY{n}{my\PYZus{}tf\PYZus{}train\PYZus{}dataset} \PY{o}{=} \PY{n}{TfDataset}\PY{p}{(}\PY{n}{learning\PYZus{}set\PYZus{}creator}\PY{p}{,} \PY{n}{my\PYZus{}train\PYZus{}sampler}\PY{p}{,} \PY{n}{batch\PYZus{}size}\PY{o}{=}\PY{l+m+mi}{16}\PY{p}{,} \PY{n}{shuffle\PYZus{}samples}\PY{o}{=}\PY{l+m+mi}{64}\PY{p}{,} \PY{n}{prefetch\PYZus{}size}\PY{o}{=}\PY{l+m+mi}{64}\PY{p}{)}
\PY{n}{count} \PY{o}{=} \PY{l+m+mi}{0}
\PY{n}{prev\PYZus{}time} \PY{o}{=} \PY{n}{time}\PY{o}{.}\PY{n}{time}\PY{p}{(}\PY{p}{)}
\PY{k}{for} \PY{n}{element} \PY{o+ow}{in} \PY{n}{my\PYZus{}tf\PYZus{}train\PYZus{}dataset}\PY{p}{:}
    \PY{n+nb}{print}\PY{p}{(}\PY{l+s+s2}{\PYZdq{}}\PY{l+s+s2}{Generation took }\PY{l+s+si}{\PYZob{}:.2f\PYZcb{}}\PY{l+s+s2}{ s}\PY{l+s+s2}{\PYZdq{}}\PY{o}{.}\PY{n}{format}\PY{p}{(}\PY{n}{time}\PY{o}{.}\PY{n}{time}\PY{p}{(}\PY{p}{)} \PY{o}{\PYZhy{}} \PY{n}{prev\PYZus{}time}\PY{p}{)}\PY{p}{,} \PY{p}{[}\PY{p}{[}\PY{n}{sample}\PY{o}{.}\PY{n}{shape} \PY{k}{for} \PY{n}{sample} \PY{o+ow}{in} \PY{n}{transformer\PYZus{}output}\PY{p}{]} \PY{k}{for} \PY{n}{transformer\PYZus{}output} \PY{o+ow}{in} \PY{n}{element}\PY{p}{]}\PY{p}{)}
    \PY{n}{count} \PY{o}{+}\PY{o}{=} \PY{l+m+mi}{1}
    \PY{n}{prev\PYZus{}time} \PY{o}{=} \PY{n}{time}\PY{o}{.}\PY{n}{time}\PY{p}{(}\PY{p}{)}

\PY{n+nb}{print}\PY{p}{(}\PY{l+s+sa}{f}\PY{l+s+s2}{\PYZdq{}}\PY{l+s+s2}{The entire my\PYZus{}tf\PYZus{}train\PYZus{}dataset contains }\PY{l+s+si}{\PYZob{}}\PY{n}{count}\PY{l+s+si}{\PYZcb{}}\PY{l+s+s2}{ elements}\PY{l+s+s2}{\PYZdq{}}\PY{p}{)}
\end{Verbatim}
\end{tcolorbox}

    It is possible to use the native Keras routines from this point onwards
to train the \gls{cnn}.

    \begin{tcolorbox}[breakable, size=fbox, boxrule=1pt, pad at break*=1mm,colback=cellbackground, colframe=cellborder]
\prompt{In}{incolor}{ }{\boxspacing}
\begin{Verbatim}[commandchars=\\\{\}]
\PY{k+kn}{from} \PY{n+nn}{deepvoxnet2}\PY{n+nn}{.}\PY{n+nn}{keras}\PY{n+nn}{.}\PY{n+nn}{optimizers} \PY{k+kn}{import} \PY{n}{SGD}
\PY{k+kn}{from} \PY{n+nn}{deepvoxnet2}\PY{n+nn}{.}\PY{n+nn}{keras}\PY{n+nn}{.}\PY{n+nn}{losses} \PY{k+kn}{import} \PY{n}{get\PYZus{}loss}

\PY{n}{my\PYZus{}own\PYZus{}unet\PYZus{}model}\PY{o}{.}\PY{n}{compile}\PY{p}{(}\PY{n}{optimizer}\PY{o}{=}\PY{n}{SGD}\PY{p}{(}\PY{n}{learning\PYZus{}rate}\PY{o}{=}\PY{l+m+mf}{1e\PYZhy{}3}\PY{p}{,} \PY{n}{momentum}\PY{o}{=}\PY{l+m+mf}{0.9}\PY{p}{,} \PY{n}{nesterov}\PY{o}{=}\PY{k+kc}{True}\PY{p}{)}\PY{p}{,} \PY{n}{loss}\PY{o}{=}\PY{n}{get\PYZus{}loss}\PY{p}{(}\PY{l+s+s2}{\PYZdq{}}\PY{l+s+s2}{dice\PYZus{}loss}\PY{l+s+s2}{\PYZdq{}}\PY{p}{,} \PY{n}{reduce\PYZus{}along\PYZus{}batch}\PY{o}{=}\PY{k+kc}{True}\PY{p}{)}\PY{p}{)}
\PY{n}{my\PYZus{}own\PYZus{}unet\PYZus{}model}\PY{o}{.}\PY{n}{fit}\PY{p}{(}\PY{n}{my\PYZus{}tf\PYZus{}train\PYZus{}dataset}\PY{p}{,} \PY{n}{epochs}\PY{o}{=}\PY{l+m+mi}{10}\PY{p}{)}
\PY{n}{visualize\PYZus{}creator\PYZus{}outputs}\PY{p}{(}\PY{n}{x\PYZus{}put\PYZus{}creator}\PY{p}{,} \PY{n}{my\PYZus{}val\PYZus{}sampler}\PY{p}{[}\PY{l+m+mi}{1}\PY{p}{]}\PY{p}{)}
\end{Verbatim}
\end{tcolorbox}

    However, how to now have your creator for training, predictions, etc., in
one place? We created the DvnModel for simplicity and to
allow for a more holistic way of defining and training or testing the
complete pipeline.

\subsection{The DvnModel object}\label{the-dvnmodel-object}

It should be clear that the set of data pairs used for training differs from the set we are interested in, 
represented by the \emph{x\_put\_creator}. However, this may not fully
suffice, e.g.~we do not yet have a prediction for our entire image. This
is because usually, even more different sets of pairs are
used. In order to group an arbitrary number of data pairs, together with
training information such as loss functions and many more, we first
create the most holistic Transformer pipeline and then create a
DvnModel.

    \begin{tcolorbox}[breakable, size=fbox, boxrule=1pt, pad at break*=1mm,colback=cellbackground, colframe=cellborder]
\prompt{In}{incolor}{ }{\boxspacing}
\begin{Verbatim}[commandchars=\\\{\}]
\PY{k+kn}{from} \PY{n+nn}{deepvoxnet2}\PY{n+nn}{.}\PY{n+nn}{components}\PY{n+nn}{.}\PY{n+nn}{model} \PY{k+kn}{import} \PY{n}{DvnModel}

\PY{c+c1}{\PYZsh{} training part of the Transformer network}
\PY{n}{x\PYZus{}affine}\PY{p}{,} \PY{n}{y\PYZus{}affine} \PY{o}{=} \PY{n}{AffineDeformation}\PY{p}{(}\PY{n}{x}\PY{p}{,} \PY{n}{rotation\PYZus{}window\PYZus{}width}\PY{o}{=}\PY{p}{(}\PY{l+m+mi}{1}\PY{p}{,} \PY{l+m+mi}{0}\PY{p}{,} \PY{l+m+mi}{0}\PY{p}{)}\PY{p}{,} \PY{n}{translation\PYZus{}window\PYZus{}width}\PY{o}{=}\PY{p}{(}\PY{l+m+mi}{10}\PY{p}{,} \PY{l+m+mi}{10}\PY{p}{,} \PY{l+m+mi}{0}\PY{p}{)}\PY{p}{)}\PY{p}{(}\PY{n}{x\PYZus{}input}\PY{p}{,} \PY{n}{y\PYZus{}input}\PY{p}{)}
\PY{n}{x\PYZus{}flip}\PY{p}{,} \PY{n}{y\PYZus{}flip} \PY{o}{=} \PY{n}{Flip}\PY{p}{(}\PY{n}{flip\PYZus{}probabilities}\PY{o}{=}\PY{p}{(}\PY{l+m+mf}{0.5}\PY{p}{,} \PY{l+m+mi}{0}\PY{p}{,} \PY{l+m+mi}{0}\PY{p}{)}\PY{p}{,} \PY{n}{n}\PY{o}{=}\PY{l+m+mi}{2}\PY{p}{)}\PY{p}{(}\PY{n}{x\PYZus{}affine}\PY{p}{,} \PY{n}{y\PYZus{}affine}\PY{p}{)}
\PY{n}{mask\PYZus{}flip} \PY{o}{=} \PY{n}{Threshold}\PY{p}{(}\PY{n}{lower\PYZus{}threshold}\PY{o}{=}\PY{l+m+mi}{0}\PY{p}{)}\PY{p}{(}\PY{n}{x\PYZus{}flip}\PY{p}{)}
\PY{n}{x\PYZus{}flip\PYZus{}crop}\PY{p}{,} \PY{n}{y\PYZus{}flip\PYZus{}crop} \PY{o}{=} \PY{n}{RandomCrop}\PY{p}{(}\PY{n}{mask\PYZus{}flip}\PY{p}{,} \PY{p}{[}\PY{p}{(}\PY{l+m+mi}{85}\PY{p}{,} \PY{l+m+mi}{85}\PY{p}{,} \PY{l+m+mi}{85}\PY{p}{)}\PY{p}{,} \PY{p}{(}\PY{l+m+mi}{53}\PY{p}{,} \PY{l+m+mi}{53}\PY{p}{,} \PY{l+m+mi}{53}\PY{p}{)}\PY{p}{]}\PY{p}{,} \PY{n}{nonzero}\PY{o}{=}\PY{k+kc}{True}\PY{p}{,} \PY{n}{n}\PY{o}{=}\PY{l+m+mi}{4}\PY{p}{)}\PY{p}{(}\PY{n}{x\PYZus{}flip}\PY{p}{,} \PY{n}{y\PYZus{}flip}\PY{p}{)}  \PY{c+c1}{\PYZsh{} notice how we can take differently sized crops around the same coordinate}
\PY{n}{y\PYZus{}train\PYZus{}pred}\PY{p}{,} \PY{n}{y\PYZus{}train} \PY{o}{=} \PY{n}{KerasModel}\PY{p}{(}\PY{n}{my\PYZus{}own\PYZus{}unet\PYZus{}model}\PY{p}{)}\PY{p}{(}\PY{n}{x\PYZus{}flip\PYZus{}crop}\PY{p}{)}\PY{p}{,} \PY{n}{y\PYZus{}flip\PYZus{}crop}
\PY{c+c1}{\PYZsh{} validation and testing part of the Transformer network}
\PY{n}{mask} \PY{o}{=} \PY{n}{Threshold}\PY{p}{(}\PY{n}{lower\PYZus{}threshold}\PY{o}{=}\PY{l+m+mi}{0}\PY{p}{)}\PY{p}{(}\PY{n}{x\PYZus{}input}\PY{p}{)}
\PY{n}{x\PYZus{}crop}\PY{p}{,} \PY{n}{y\PYZus{}crop} \PY{o}{=} \PY{n}{RandomCrop}\PY{p}{(}\PY{n}{mask}\PY{p}{,} \PY{p}{[}\PY{p}{(}\PY{l+m+mi}{85}\PY{p}{,} \PY{l+m+mi}{85}\PY{p}{,} \PY{l+m+mi}{85}\PY{p}{)}\PY{p}{,} \PY{p}{(}\PY{l+m+mi}{53}\PY{p}{,} \PY{l+m+mi}{53}\PY{p}{,} \PY{l+m+mi}{53}\PY{p}{)}\PY{p}{]}\PY{p}{,} \PY{n}{nonzero}\PY{o}{=}\PY{k+kc}{True}\PY{p}{,} \PY{n}{n}\PY{o}{=}\PY{l+m+mi}{100}\PY{p}{)}\PY{p}{(}\PY{n}{x\PYZus{}input}\PY{p}{,} \PY{n}{y\PYZus{}input}\PY{p}{)}  \PY{c+c1}{\PYZsh{} notice that n is much larger now}
\PY{n}{y\PYZus{}val\PYZus{}pred}\PY{p}{,} \PY{n}{y\PYZus{}val} \PY{o}{=} \PY{n}{KerasModel}\PY{p}{(}\PY{n}{my\PYZus{}own\PYZus{}unet\PYZus{}model}\PY{p}{)}\PY{p}{(}\PY{n}{x\PYZus{}crop}\PY{p}{)}\PY{p}{,} \PY{n}{y\PYZus{}crop}
\PY{n}{y\PYZus{}val\PYZus{}pred\PYZus{}buffer} \PY{o}{=} \PY{n}{Buffer}\PY{p}{(}\PY{n}{buffer\PYZus{}size}\PY{o}{=}\PY{k+kc}{None}\PY{p}{)}\PY{p}{(}\PY{n}{y\PYZus{}val\PYZus{}pred}\PY{p}{)}
\PY{n}{y\PYZus{}full\PYZus{}val\PYZus{}pred}\PY{p}{,} \PY{n}{y\PYZus{}full\PYZus{}val} \PY{o}{=} \PY{n}{Put}\PY{p}{(}\PY{n}{reference\PYZus{}connection}\PY{o}{=}\PY{n}{x\PYZus{}input}\PY{p}{)}\PY{p}{(}\PY{n}{y\PYZus{}val\PYZus{}pred\PYZus{}buffer}\PY{p}{)}\PY{p}{,} \PY{n}{y\PYZus{}input}
\PY{c+c1}{\PYZsh{} creating a single DvnModel of the complete pipeline, typically in a [y\PYZus{}pred, y\PYZus{}true, sample\PYZus{}weights] order, similar to keras, but substituted x by y\PYZus{}pred}
\PY{n}{my\PYZus{}super\PYZus{}cool\PYZus{}dvnmodel} \PY{o}{=} \PY{n}{DvnModel}\PY{p}{(}
    \PY{n}{outputs}\PY{o}{=}\PY{p}{\PYZob{}}
        \PY{l+s+s2}{\PYZdq{}}\PY{l+s+s2}{train}\PY{l+s+s2}{\PYZdq{}}\PY{p}{:} \PY{p}{[}\PY{n}{y\PYZus{}train\PYZus{}pred}\PY{p}{,} \PY{n}{y\PYZus{}train}\PY{p}{]}\PY{p}{,}
        \PY{l+s+s2}{\PYZdq{}}\PY{l+s+s2}{full\PYZus{}val}\PY{l+s+s2}{\PYZdq{}}\PY{p}{:} \PY{p}{[}\PY{n}{y\PYZus{}full\PYZus{}val\PYZus{}pred}\PY{p}{,} \PY{n}{y\PYZus{}full\PYZus{}val}\PY{p}{]}\PY{p}{,}
        \PY{l+s+s2}{\PYZdq{}}\PY{l+s+s2}{full\PYZus{}test}\PY{l+s+s2}{\PYZdq{}}\PY{p}{:} \PY{p}{[}\PY{n}{y\PYZus{}full\PYZus{}val\PYZus{}pred}\PY{p}{]}
    \PY{p}{\PYZcb{}}
\PY{p}{)}
\end{Verbatim}
\end{tcolorbox}

    The DvnModel can be used to call the training routines directly, without
the need to create Creator objects and TfDataset objects. Furthermore,
the DvnModel can be saved, storing all the Creator objects
and the KerasModel. We advise users to have a careful look at the
DvnModel class for all its use cases and how \texttt{compile},
\texttt{fit}, \texttt{predict} methods, etc, can be used similar to the
Keras API.

    \section{Other functionalities in DVN2}\label{other-functionalities-in-dvn2}

The main contribution of \gls{dvn2} is the intuitive way of
implementing preprocessing, sampling and postprocessing and providing a
way to store everything under one single object. However, there are
other unexplored fruits available.

\subsection{Utilities}\label{utilities}

Under \texttt{deepvoxnet.utilities} there are many useful utility
functions available:

\begin{itemize}
\item
  \textbf{conversions.py} Here, you can find many useful loading and
  conversion functions. For example, reading a Dicom file (including
  time series such as \gls{ctp} data) from disk, conversions between
  Nifti, SimpleITK and Dicom.
\item
  \textbf{transformations.py} Perfect for resampling,
  registration, etc., mainly starting from Nifti files. This is unlike
  many other libraries that require SimpleITK file formats or you need
  to use via the Terminal.
\item
  \textbf{drawing.py} Some cool functions to draw overlays.
\end{itemize}

\subsection{Analysis}\label{analysis}

Under \texttt{deepvoxnet.analysis} we can find several handy tools to:

\begin{itemize}
\item
  \textbf{data.py} The Data class can be viewed as a single-column
  Pandas Dataframe with additional functionalities, specifically to work
  with data that has MultiIndex structure as
  Dataset~\textgreater~Case~\textgreater~Record. We could group or compute
  statistics across either of these indices.
\item
  \textbf{analysis.py} The Analysis class can be viewed as a
  multi-column Pandas Dataframe with additional functionalities. For
  example, when having two columns, we can apply a metric function to
  each row such that a Data object is returned.
\item
  \textbf{plotting.py} This library uses the Figure class for plotting
  such that you can draw figures using an absolute scale. Furthermore,
  since typical plotting libraries cannot handle NaNs, we have the
  Series, SeriesGroup and GroupedSeries objects that compute statistics
  and can be used to plot, even if they contain NaNs.
\end{itemize}

    \section{Conclusion}\label{conclusion}

We have provided a novel \gls{cnn} framework that can be used to design
end-to-end \gls{cnn} pipelines for either 1D, 2D or 3D segmentation or
classification. Some of the key design aspects of \gls{dvn2} are that it stays
close to the supervised learning paradigm and keeps track of the
spatial aspects of data pairs. While potentially memory and computationally
hungry, \gls{dvn2} provides users with a multitude of intuitive abstractions
meant to increase the awareness of some of the underlying mechanisms of
\gls{cnn}s. On top of this, \gls{dvn2} provides other useful utility and analysis
functions for medical image analysis, such as functions related
to image I/O, registration, statistical analysis and visualization.

\bibliographystyle{plain}
\bibliography{references}

\end{document}